\documentclass{article}

\usepackage{microtype}
\usepackage{graphicx}
\usepackage{subfigure}
\usepackage{booktabs} 
\usepackage{amsfonts}
\usepackage{amsmath}
\usepackage{colortbl}
\usepackage[table,xcdraw]{xcolor}
\usepackage{multirow}
\definecolor{darkgreen}{RGB}{0,100,0}
\usepackage{hyperref}
\usepackage{tcolorbox}

\usepackage[accepted]{mlsys2026}
\usepackage{eso-pic}
\usepackage{graphicx}

\mlsystitlerunning{SkipKV: Selective Skipping of KV Generation and Storage for Efficient Inference with Large Reasoning Models}

\usepackage[firstpage]{draftwatermark}
\SetWatermarkAngle{0}
\SetWatermarkText{%
  \hspace*{\dimexpr\textwidth-1.5cm\relax}%
  \raisebox{25cm}{%
    \includegraphics[width=1.2cm]{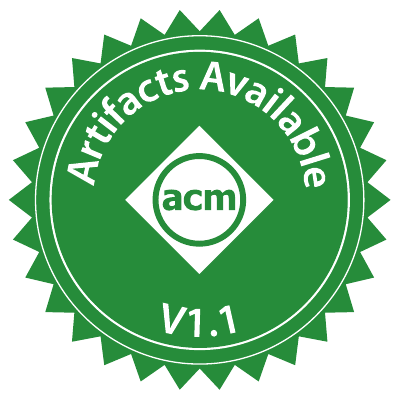}%
    \hspace{2pt}%
    \includegraphics[width=1.2cm]{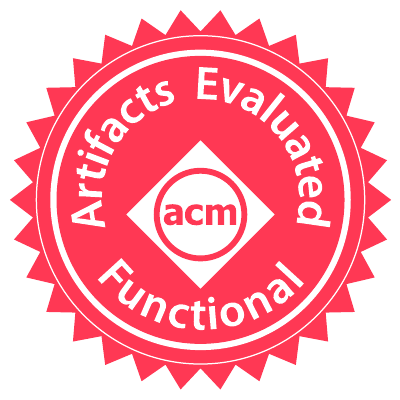}%
  }%
}

\begin{document}

\twocolumn[
\mlsystitle{SkipKV: Selective Skipping of KV Generation and Storage for Efficient Inference with Large Reasoning Models}




\setcounter{footnote}{0}
\renewcommand{\thefootnote}{\fnsymbol{footnote}}

\begin{mlsysauthorlist}
\mlsysauthor{Jiayi Tian\thanks{Initial ideation and majority of the work was done during Jiayi's internship at Intel.}}{ucsb,intel}
\mlsysauthor{Seyedarmin Azizi}{usc}
\mlsysauthor{Yequan Zhao}{ucsb}
\mlsysauthor{Erfan Baghaei Potraghloo}{usc}
\mlsysauthor{Sean McPherson}{intel}
\mlsysauthor{Sharath Nittur Sridhar}{intel}
\mlsysauthor{Zhengyang Wang}{ucsb}
\mlsysauthor{Zheng Zhang}{ucsb}
\mlsysauthor{Massoud Pedram}{usc}
\mlsysauthor{Souvik Kundu}{intel}
\end{mlsysauthorlist}
\mlsysaffiliation{ucsb}{University of California, Santa Barbara}
\mlsysaffiliation{intel}{Intel Labs}
\mlsysaffiliation{usc}{University of Southern California}


\mlsyscorrespondingauthor{Souvik Kundu}{souvikk.kundu@intel.com}

\mlsyskeywords{Machine Learning, MLSys}

\vskip 0.3in

\begin{abstract}
Large reasoning models (LRMs) often incur significant key-value (KV) cache overhead, due to their linear growth
with the verbose chain-of-thought (CoT) reasoning. 
This incurs both memory overhead and throughput bottlenecks, limiting efficient deployment.
To reduce KV cache size during inference, we first investigate the effectiveness of existing KV cache eviction methods for CoT reasoning. 
Interestingly, we find that due to unstable token-wise scoring and reduced effective KV budget caused by padding, state-of-the-art (SoTA) eviction methods fail to maintain accuracy in multi-batch settings.
Additionally, these methods often generate longer sequences than the original model without eviction, as semantic-unaware token-wise eviction leads to repeated revalidation during reasoning.
To address these issues, we present \textbf{SkipKV}, a \textbf{\textit{training-free}} KV compression method that performs selective \textit{eviction} and \textit{generation}, operating at a coarse-grained, sentence-level sequence removal for efficient CoT reasoning. In specific, it introduces a \textit{sentence-scoring metric} to identify and remove highly similar sentences while maintaining semantic coherence. 
To suppress redundant generation, SkipKV dynamically adjusts a steering vector to update the hidden activation states during inference, enforcing the LRM to generate concise responses.
Extensive evaluations on multiple reasoning benchmarks demonstrate that SkipKV achieves up to $\mathbf{26.7}\%$ higher accuracy compared to baseline methods, at a similar compression budget. Additionally, compared to SoTA, SkipKV yields up to $\mathbf{1.6}\times$ shorter generation length while improving throughput by up to $\mathbf{1.7}\times$. 
Our code is released at: \href{https://github.com/TTTTTTris/SkipKV}{https://github.com/TTTTTTris/SkipKV}.

\end{abstract}
]



\printAffiliationsAndNotice{}  

\section{Introduction}
Large language models (LLMs) have demonstrated remarkable capabilities in complex reasoning with advancements enabling them to perform multi-step mathematical derivations \cite{trinh2024solving, luowizardmath, ahn2024large} and code generation \cite{wang2023review, zhong2024can, mahbub2025prism}. However, reasoning-oriented models (e.g., DeepSeek-R1 \cite{guo2025deepseek}) face a critical deployment bottleneck: their tendency to generate lengthy and often redundant reasoning traces that lead to unsustainable memory demands \cite{chen2025do}.
In particular, the large token count linearly increases the key-value (KV) cache memory of the autoregressive large reasoning models (LRMs), often making them the dominant component for large reasoning depth.
For instance, a DeepSeek-R1-Distill-Llama-8B model may generate over 32K tokens when solving a single complex math problem; for a batch size of 10, it needs $\mathord{\sim}2.5\times$ larger KV cache memory compared to the model weights. 
In addition to the significant memory overhead, the growing KV cache impacts the throughput of the memory-bound decoding stage of LRMs. This highlights the need for reasoning KV cache compression an important research paradigm for long chain-of-thought (CoT) reasoning.
\begin{figure}[!t]
    \centering
    \includegraphics[width=0.9\linewidth]{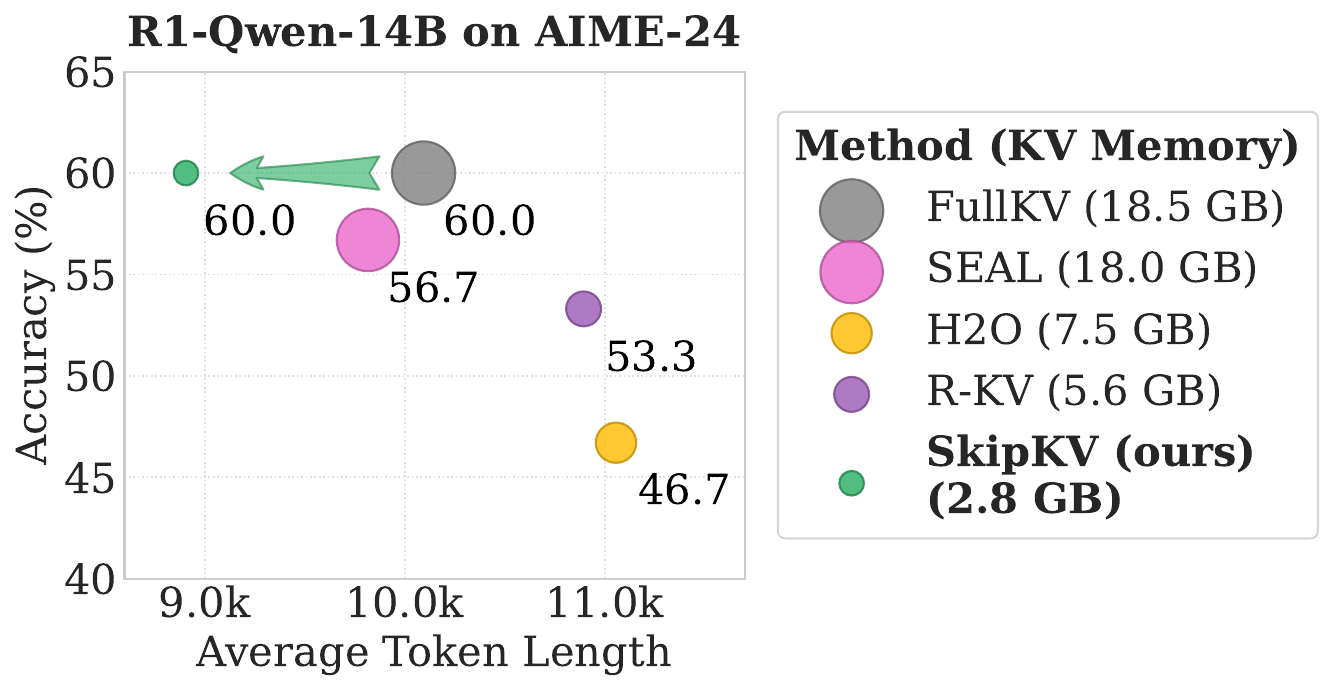}
    \vspace{-3mm}
    \caption{Comparison of KV cache eviction methods. Marker size denotes KV memory usage. SkipKV yields shorter generation length while maintaining high accuracy under a smaller KV budget.}
    \vspace{-5mm}
    \label{fig:intro_result}
\end{figure}
\begin{figure*}[!ht]
    \centering
    \includegraphics[width=0.90\linewidth]{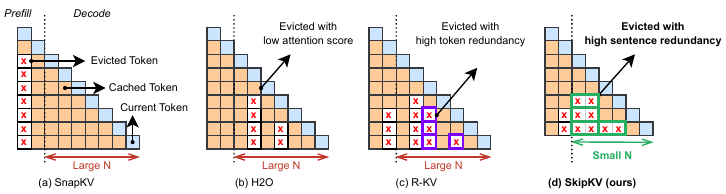}
    \vspace{-3mm}
    \caption{
    Comparison of KV cache eviction methods during token generation.
    Cached tokens marked with {\color{red}${\times}$} indicate evicted positions. 
    (a) SnapKV performs one-time eviction after prefill; 
    (b) H2O evicts tokens with low cumulative attention scores; 
    (c) R-KV prunes redundant tokens based on token-level similarity (purple);
    (d) SkipKV (ours) groups tokens within sentences (green) to evict high sentence-redundancy regions, achieving high accuracy and shorter generation length ($N$). 
    }
    \vspace{-3mm}
    \label{fig:intro}
\end{figure*}

Despite significant research, most of the existing KV compression methods \cite{zhang2023h2o, xiaoefficient, li2024snapkv, tangquest} target the KV compression of long-context prefill and lose their efficacy in long CoT tasks for LRMs. For example, H2O~\cite{zhang2023h2o} improves decoding efficiency by retaining a compact subset of KV pairs based on cumulative historical attention scores. Yet, its relies solely on attention magnitude and overlooks semantic coherence across multi-step reasoning spans. SnapKV~\cite{li2024snapkv} leverages attention-based importance estimation to compress the KV cache during the prefill phase, achieving strong performance in long-context scenarios but shows a significant accuracy drop for CoT reasoning tasks. Only recently R-KV~\cite{cai2025r} highlighted a potential limitation of these works, by focusing on \textit{repetitive and redundant} CoT tokens. It then proposed a redundancy-aware metric to effectively remove repetitive tokens along the reasoning path, achieving over 80\% KV-cache compression while maintaining strong accuracy. 
However, despite good accuracy on single-batch settings, R-KV often suffers from a significant accuracy drop for multi-batch settings, limiting its scalable deployment.
Moreover, its token-level eviction granularity disregards higher-level semantic structure, often resulting in unnecessarily prolonged reasoning paths.
Fig.~\ref{fig:intro} compares representative KV-cache eviction strategies. Here, we focus on permanent eviction methods, where evicted tokens are no longer accessible in subsequent decoding steps, thereby yielding genuine memory savings.

\textbf{Our Contributions.} To address these limitations, we first investigate the limitations of existing token eviction methods. 
Our analysis reveals that incoherent token eviction can result in unstable reasoning with reduced accuracy and prolonged generation. 
Specifically, we find that the fragmented token histories often induce overthinking and force the CoT to generate more tokens for a fixed KV budget. 
Based on these insights, we then propose \textbf{SkipKV}, a sentence-aware selective KV eviction and generation framework designed for efficient CoT reasoning. 
SkipKV not only preserves reasoning coherence but also \textit{achieves a better trade-off between accuracy and generation length at a fixed KV cache memory budget} (Fig. \ref{fig:intro_result}). At its core, SkipKV consists of two key components, namely the \textbf{sentence-primary scoring method} and the \textbf{adaptive steering method} to efficiently skip token storage and generation, respectively. Specifically, it relies on a \textit{sentence-primary scoring metric} that enables selective KV storage skipping while preserving the semantic coherence of the reasoning process. The \textbf{adaptive steering mechanism} suppresses uninformative or redundant sentences during the generation. 
Additionally, SkipKV adopts a \textbf{batch grouping policy} that reduces the number of padding tokens, thereby freeing KV-cache space for valid tokens and improving stability and consistency in multi-batch reasoning with KV eviction.

To demonstrate the effectiveness of SkipKV, we conduct comprehensive evaluations on DeepSeek-R1-Qwen-7B, R1-Qwen-14B, and R1-Llama-8B LRMs across four reasoning benchmarks: AIME-24, LiveCodeBench, MATH-500, and GSM8K. 
As shown in Fig.~\ref{fig:intro_result}, SkipKV outperforms state-of-the-art (SoTA) methods, achieving $\mathbf{6.7}\%$ higher accuracy and $\mathbf{22}\%$ shorter generation lengths, with $\mathbf{2}\times$ KV memory compression on R1-Qwen-14B evaluated on AIME-24.


\begin{figure*}[!t]
    \centering
    \includegraphics[width=.95\linewidth]{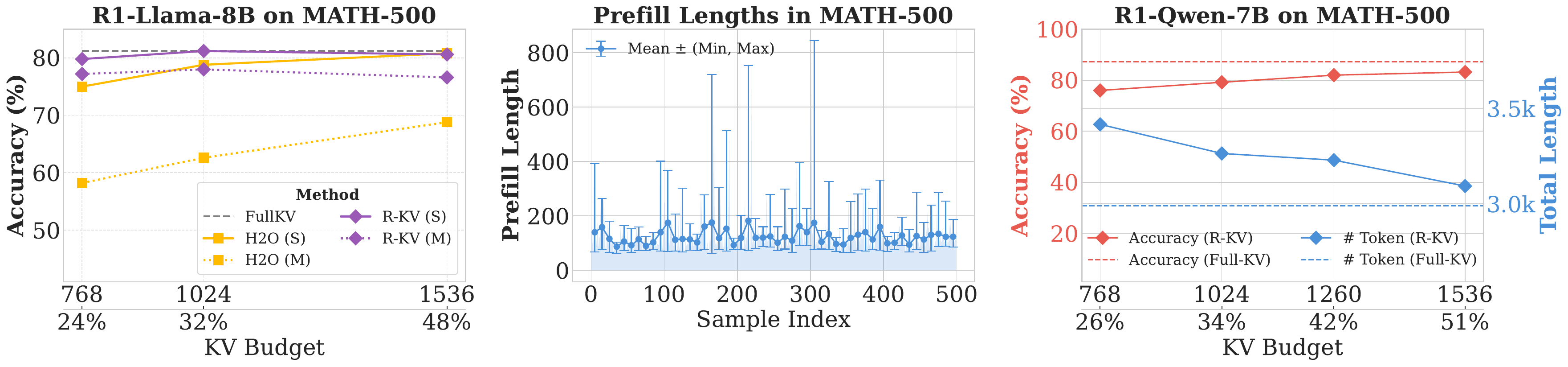}
    \vspace{-3mm}
    \caption{
        \textbf{Left:} Accuracy comparison for single- and multi-batch decoding of H2O \cite{zhang2023h2o} and R-KV \cite{cai2025r}. 
        \textbf{Center:} Visualization of the prefill token length distribution of MATH-500, and the min-max range of each batch (batch size, bs = 10). 
        \textbf{Right:} Accuracy and generated token length versus KV budget with R-KV eviction on MATH-500 (bs = 10).
    }
    \vspace{-3mm}
    \label{fig:motivation}
\end{figure*}
\section{Related Works}


\textbf{KV Cache Compression Methods.}
To mitigate the growing KV cache memory footprint, existing research on \textit{inference-time} KV cache compression can be broadly classified into two categories: KV cache \textit{eviction} \cite{zhang2023h2o, li2024snapkv, behnamrocketkv, tangquest} and \textit{quantization} \cite{liu2024kivi, kang2024gear, ramachandran2025thinkv}. 

The KV eviction methods that \textit{permanently remove redundant tokens}, including H2O \cite{zhang2023h2o} and SnapKV \cite{li2024snapkv}, primarily rely on the token importance ranked based on attention scores to remove low-scoring KV tokens and meet a fixed KV budget, reducing the total memory demand. 
Chunk-based eviction methods, such as ChunkKV \cite{liu2025chunkkv}, extend this idea by grouping tokens into chunks and performing eviction at the chunk level, which reduces eviction overhead and improves memory locality while still relying on attention-derived importance signals.
These works, while performing well on non-reasoning benchmarks with large prefill length, fail to demonstrate good accuracy for CoT compression. Only recently, a contemporary work, R-KV \cite{cai2025r} identifies a key limitation of these methods: \textit{not adhering to the redundancy in the token importance scoring}. R-KV proposes a redundancy-aware token scoring to yield SoTA compression-accuracy trade-off for CoT reasoning. However, R-KV still suffers from a larger generation length and reduced accuracy in multi-batch decoding. 

Other selection-based eviction methods, such as Quest \cite{tangquest}, keep FullKV in global memory and bring a fractional chunk into local memory that remains relevant to the query. 
While such approaches can preserve accuracy by maintaining access to the complete historical context, the KV cache still grows linearly with sequence length, which can limit decoding throughput, reduce batch size scalability, and lead to out-of-memory issues for long chain-of-thought (CoT) reasoning workloads \cite{liu2025freekv, hu2025raas}.

In contrast to eviction, KV quantization reduces memory footprint by storing KV tokens at lower precision. However, for large reasoning models (LRMs), quantization may degrade CoT reasoning performance \cite{liu2025quantization}. 
In this work, we focus on improving KV eviction, while KV quantization remains an orthogonal direction.

\textbf{Other KV Cache Reduction Methods.}
Apart from compression, earlier works have explored approaches to reduce the KV memory footprint via different methods, including KV \textit{sharing}, \textit{early exiting}, and \textit{steering}.
Sharing methods, like KVsharer \cite{yang2024kvsharer}, relies on tensor similarity between KV caches of different layers of a model, to allow them to be shared across $\geq 1$ layer. 
However, this sharing costs a significant accuracy drop on CoT reasoning tasks. 
DEER~\cite{yang2025dynamic} monitors transition tokens during the reasoning process and estimates the confidence of early termination by inducing trial answer tokens, thereby reducing reasoning length through adaptive early exits. 
KV steering \cite{chen2025seal, azizi2025activation} on the other hand introduces a latent-space calibration method to identify and manipulate non-important thoughts in CoT to reduce the final reasoning sequence length. While these methods can reduce the generation token count to improve per-sample memory, they cannot meet a fixed KV budget and their batch-level scalability remains an open problem.

\section{Motivational Case Studies}
In this section, we analyze the key limitations of the representative SoTA CoT token eviction methods.
\label{sec:problems}

\begin{center}
\vspace{-1mm}
\begin{tcolorbox}[width=0.48\textwidth]
\vspace{-2mm}
\textbf{Observation 1:} \textit{With KV eviction, reasoning accuracy drops in multi-batch decoding compared to that with single-batch.}
\vspace{-2mm}
\end{tcolorbox}
\vspace{-2mm}
\end{center}
\textbf{Description.} The benefit of KV cache eviction methods lies in their ability to reduce KV memory usage, thereby enabling larger batch size with improved generation throughput.  To analyze the accuracy robustness of the existing SoTA KV eviction methods \cite{cai2025r, zhang2023h2o} on CoT reasoning tasks, we evaluate the accuracy for a batch size of 1 and 10, respectively. Specifically, we measure the accuracy on the MATH-500 with the R1-Llama-8B at different KV budgets (as a fraction of the KV memory, with FullKV as $100\%$). As depicted in Fig.~\ref{fig:motivation} (left), the multi-batch decoding leads to substantial accuracy degradation, for both R-KV and H2O, particularly at lower KV budgets.
This limitation may be attributed to the \textit{effective KV budget reduction} for the multi-batch scenario due to the padding tokens. These tokens are added to align variable-length prefill sequences across different samples in a batch. 
Upon enabling a fixed memory budget via KV eviction, the padding tokens consume memory space resulting in a reduction in the effective KV budget, and thus potentially reduce accuracy. 
In addition, these tokens can also distort attention distributions, leading to unstable token-level importance estimation that can further contribute to reasoning consistency degradation. 
As depicted in Fig. \ref{fig:motivation} (center), the prefill token length in MATH-500 varies widely across samples, with some variations exceeding 400 tokens in a batch. This variation leads to excessive padding when sequences are batched together, increasing KV cache waste and exacerbating the performance drop in multi-batch decoding with eviction.

\begin{center}
\vspace{-1mm}
\begin{tcolorbox}[width=0.48\textwidth]
\vspace{-2mm}
\textbf{Observation 2:} \textit{At reduced KV budgets the total generation length often increases compared to models without any KV compression.}
\vspace{-2mm}
\end{tcolorbox}
\vspace{-2mm}
\end{center}
\textbf{Description.} Fig.~\ref{fig:motivation} (right) shows the accuracy and average token length (continuous lines) with DeepSeek-R1-Qwen 7B for a batch size of 10, evaluated on MATH-500. In addition to reduced accuracy, models with R-KV compression consistently generate longer sequences across all KV budgets compared to models without compression.
This observation indicates that token eviction can cause the model to extend its reasoning chain and generate longer outputs.
This may be attributed to the loss of contextual information in the KV cache: when portions of the valid context are removed, the model may compensate by regenerating reasoning steps or adding redundant segments to recover missing information.
The longer generations result in increased computational cost due to additional forward passes, thereby offsetting efficiency gains from cache size reduction.
%
\begin{figure}[!t]
    \centering
    \fcolorbox{blue!70!black}{white}{%
    \includegraphics[width=.9\linewidth]{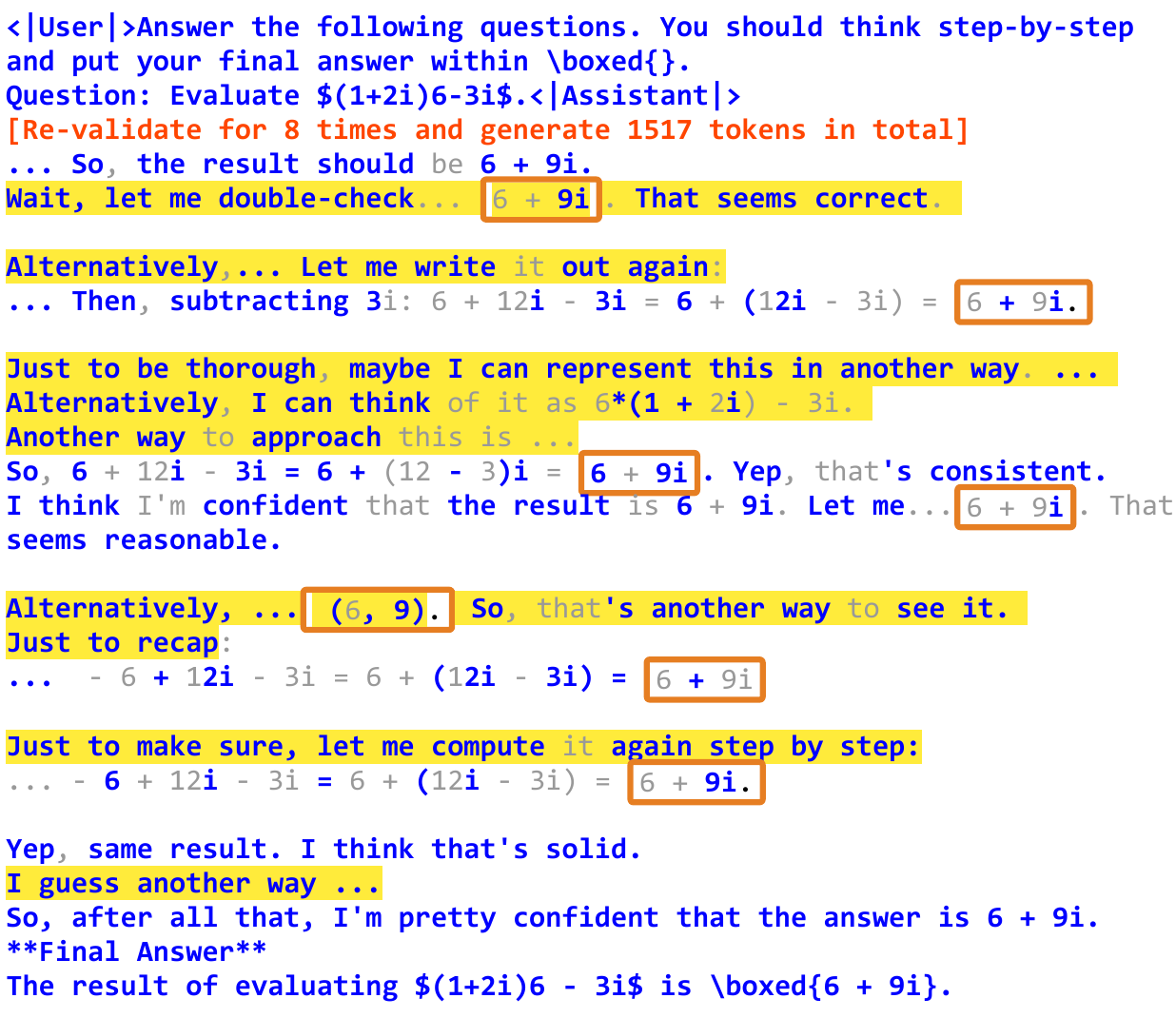}
    }
    \vspace{-2mm}
    \caption{
    SoTA token-based eviction \cite{cai2025r} often selects fragmented tokens from the final answer $(6 + 9i)$ (highlighted in orange boxes), causing repeated self-validation and non-execution thoughts (highlighted in yellow). Tokens in \textcolor{blue}{blue} are retained, while those in \textcolor{gray}{gray} are evicted.}
    \vspace{-3mm}
    \label{fig:demo}
\end{figure}
\begin{center}
\vspace{-3mm}
\begin{tcolorbox}[width=0.48\textwidth]
\vspace{-2mm}
\textbf{Observation 3:} \textit{Token-level eviction often causes fragmented removal of words, leading the LRM to overthink.}
\vspace{-2mm}
\end{tcolorbox}
\vspace{-2mm}
\end{center}

\textbf{Description.} 
To further investigate the cause of increased generation length, we visualize the tokens retained after R-KV eviction in Fig.~\ref{fig:demo}.
We select a sample from MATH-500 and show model outputs where the retained tokens are highlighted in blue and the evicted ones in gray.

The limitations here can be summarized in two ways.
First, eviction purely based on token-level redundancy scores \cite{cai2025r} often removes numbers from crucial mathematical computation steps, thereby disrupting the reasoning flow.
Second, token-level eviction may often retain fragmented tokens from the correct answer, misleading the LRM to re-validate partial results and generate unnecessarily long or uncertain reasoning chains.
For instance, in Fig.~\ref{fig:demo}, the tokens retained by R-KV frequently include disjoint numerical entities from intermediate reasoning steps (e.g., $6 + 12i - 3i \rightarrow + 2i$) or from the final answer (e.g., $(6,9) \rightarrow (,9)$).
Such fragmented retention causes the model to repeatedly re-derive or re-check parts of the answer, resulting in longer yet redundant reasoning trajectories.

These findings highlight the need for a coherent eviction policy that operates on higher-level semantic units (e.g., sentences or reasoning segments) to achieve a better trade-off between accuracy and generation length.
%
\begin{figure}[!t]
    \centering
    \includegraphics[width=0.95\linewidth]{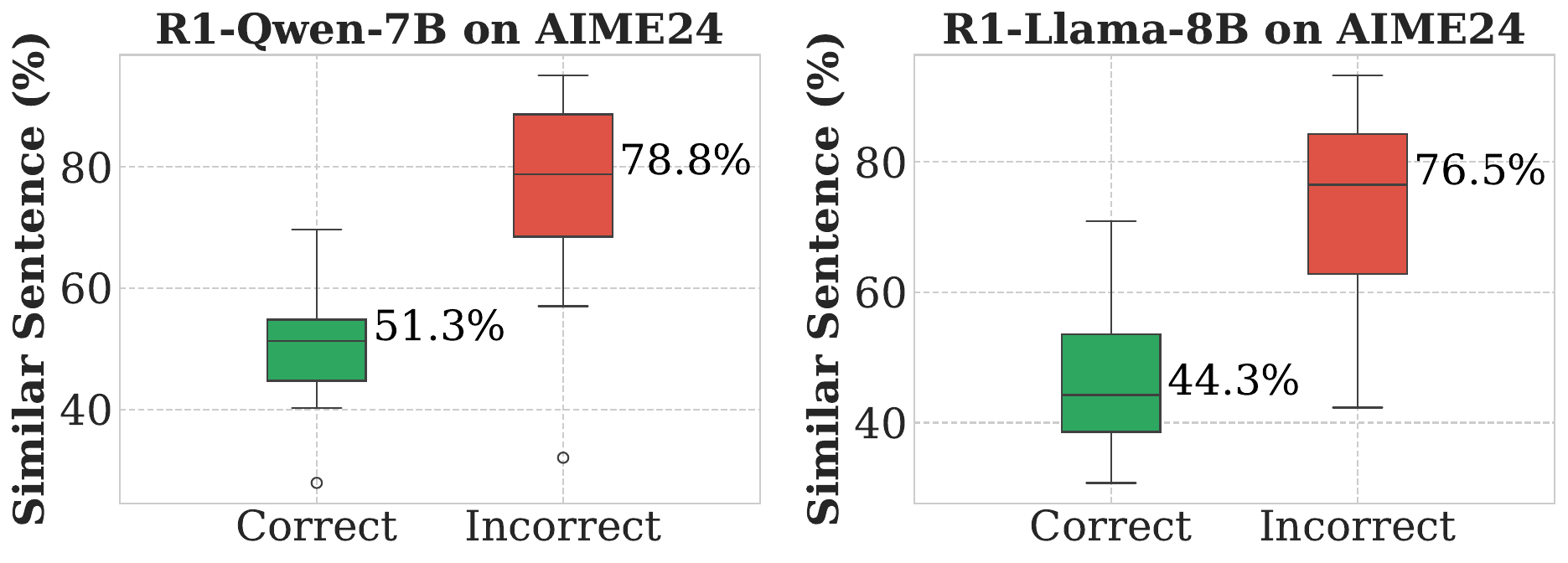}
    \vspace{-2mm}
    \includegraphics[width=0.95\linewidth]{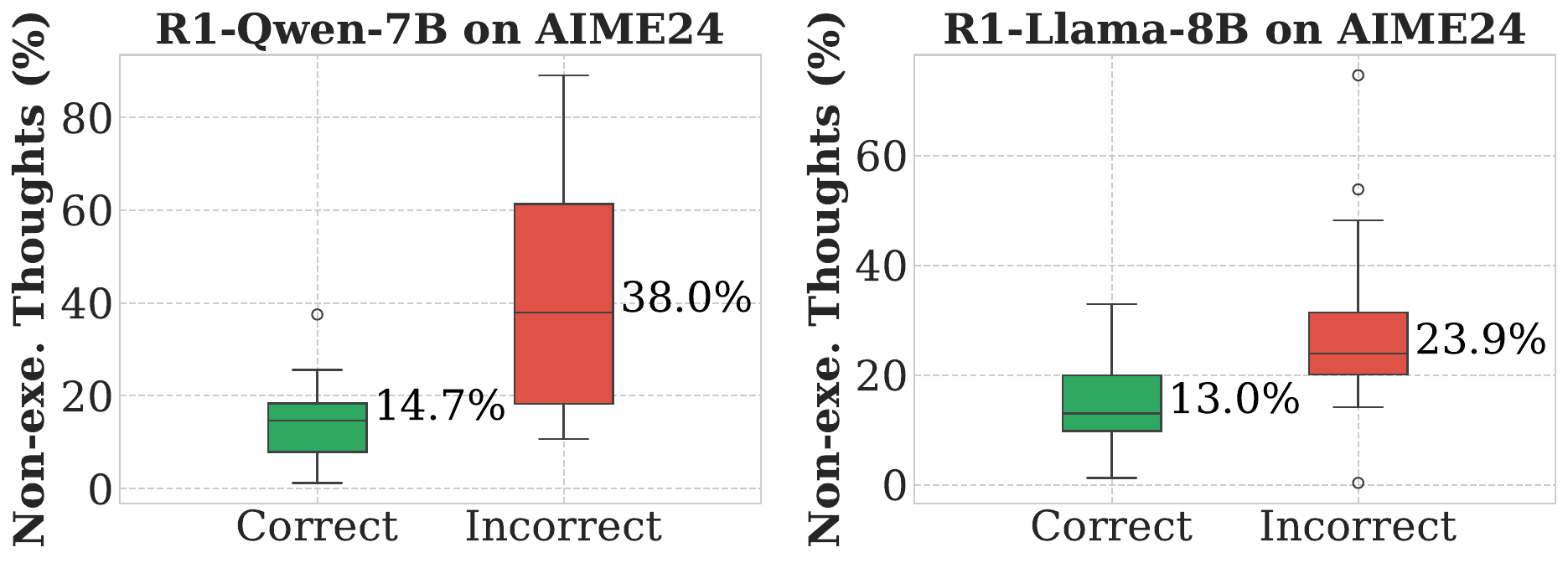}
    \caption{Statistics on the ratio of high-similarity sentences (top) and non-execution thoughts (bottom) for samples correctly and incorrectly answered by the models.}
    \vspace{-3mm}
    \label{fig:sentence}
\end{figure}
\section{SkipKV: Methodology}
\label{sec:method}
\begin{figure*}[!t]
    \centering
    \includegraphics[width=0.95\linewidth]{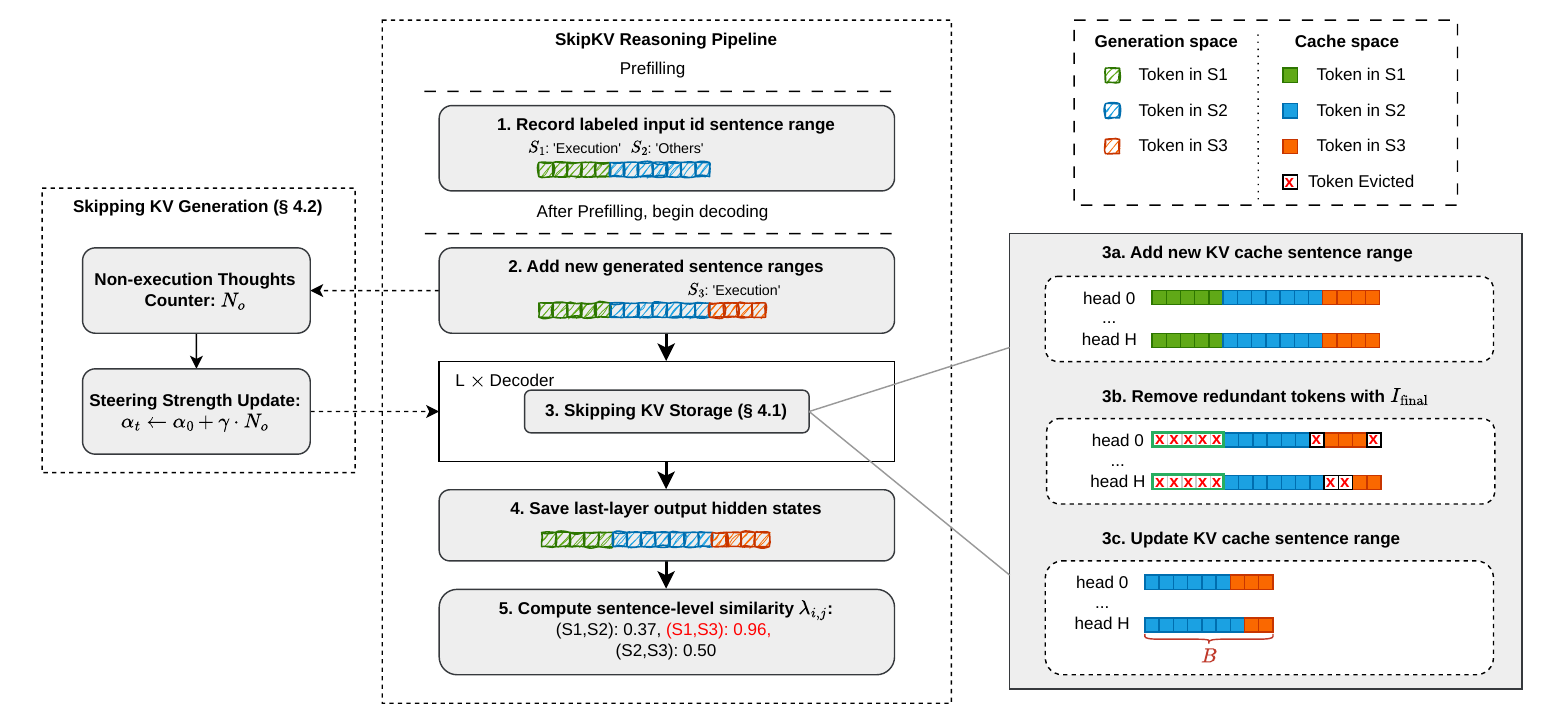}
    \vspace{-2mm}
    \caption{Overview of the SkipKV framework.
    It selectively skips KV-cache storage and generation by sentence-level redundancy detection.
    The central reasoning pipeline illustrates the end-to-end process from prefill to decoding, where input sentence ranges and types are recorded, sentences are scored, and the KV cache is compressed.
    The left panel depicts the adaptive steering mechanism used to skip KV generation ($\S$~\ref{sec:adaptive-steering}), while the right panel illustrates sentence-level KV storage skipping ($\S$~\ref{sec:sentence-scoring}), including the KV range monitoring logic (3a, 3c) and the sentence-oriented eviction strategy (3b).}
    \vspace{-3mm}
    \label{fig:overview}
\end{figure*}

In this section, we first analyze the sentence-level properties of reasoning traces motivating the design of SkipKV (illustrated in Fig.~\ref{fig:overview}).
We then present the two core components of SkipKV: (1) sentence-level skipping of the KV cache storage  guided by semantic redundancy scoring; and (2) adaptive KV steering for controlled generation to suppress unnecessary thought expansion. Finally, we present a simple yet effective batch-grouping-driven eviction strategy to enhance the accuracy robustness in multi-batch decoding. 
\textbf{Definition: Pairwise Sentence Similarity} (PSS) \textit{between two sentences is defined as the measured cosine similarity between the vector embeddings of the two. 
Let the vector embedding of two sentences be $\mathbf{v}_i, \mathbf{v}_j \in \mathbb{R}^{d}$, respectively. The PSS score is computed as}
\vspace{-2mm}
\begin{equation}
    \mathrm{PSS}(\textbf{v}_i, \textbf{v}_j) = \mathbf{v}_i^\top\mathbf{v}_j.
    \vspace{-2mm}
    \label{eqa:pss}
\end{equation}
To understand the differences between successful and failed reasoning trajectories from the semantic perspective, we perform a fine-grained analysis of sentence- (thought-) level properties in the model generated output with FullKV.
We focus on two aspects: (i) the semantic similarity among generated sentences, and (ii) the ratio of non-execution (less important) thoughts. Notably, following the definition of \cite{chen2025seal}, here we classify the generated tokens as \textbf{execution} (important tokens relevant to the actual answer) and \textbf{non-execution} (less important tokens generated to verify the response via reflection or transition) thoughts. 

\begin{center}
\vspace{-1mm}
\begin{tcolorbox}[width=0.48\textwidth]
\vspace{-2mm}
\textbf{Observation 4:} \textit{Both correct and incorrect reasoning responses generate highly similar sentences with the latter scenario usually generating higher $\%$ of similar sentences.}
\vspace{-2mm}
\end{tcolorbox}
\vspace{-2mm}
\end{center}
\textbf{Description.} We perform an experiment with R1-Qwen-7B and R1-Llama-8B on AIME24 to compute the PSS. In particular, we first collect model output texts and segment them into individual sentences separated by newline and punctuation-based patterns (e.g., `\textbackslash n\textbackslash n').
Each sentence is then encoded into a semantically meaningful embedding vector using a BERT-based sentence transformer \cite{reimers2019sentence}.
We then compute the PSS for all the sentence-embedding pairs and mark pairs with a PSS score $\geq 0.95$ as highly similar, yielding a measure of semantic redundancy. Interestingly, as shown in Fig.~\ref{fig:sentence} (top), incorrect responses consistently exhibit up to $1.7\times$ higher $\%$ of similar sentences as compared to correct responses.


\begin{center}
\vspace{-1mm}
\begin{tcolorbox}[width=0.48\textwidth]
\vspace{-2mm}
\textbf{Observation 5:} \textit{Incorrect responses generate significantly higher $\%$ of non-execution thoughts compared to the correct ones.}
\vspace{-2mm}
\end{tcolorbox}
\vspace{-2mm}
\end{center}
\textbf{Description.} 
As shown in Fig. \ref{fig:sentence} (bottom), incorrect generations consistently exhibit higher ratios of non-execution thoughts compared to correct responses.
In particular, when evaluating on AIME24 with R1-Qwen-7B and R1-Llama-8B, respectively, we observe around 
$2.6 \times$ and $1.8 \times$ higher $\%$ of non-execution thoughts in incorrect outputs compared to correct outputs.
These patterns indicate that when models fail, they tend to produce repetitive or stagnant reasoning steps, revisiting semantically similar content or generating meta-level commentary rather than performing concrete problem-solving actions.

\subsection{Skipping KV Storage with Sentence-level Scoring}
\label{sec:sentence-scoring}
Here, we first introduce the sentence-level similarity scoring and then present the cumulative scoring mechanism to decide which sentences and tokens to evict.

\textbf{Sentence Similarity Score.}
We use this score to compute the sentence-level redundancy.
However, using a sentence transformer to generate the latent vector representation of the sentences is impractical as it costs significant compute overhead during decoding.
Instead, we leverage the last-layer hidden state, denoted as $\mathbf{H} \in \mathbb{R}^{bs \times N \times d}$, as latent contextual representations of the sentence segments. Specifically, for each sample in a batch, we identify the start and end indices of each sentence segment by partitioning the total sequence ($N$) into small chunks that are separated by punctuation-based delimiters (Step 1–2 in Fig.~\ref{fig:overview}). For a sentence $i$ at batch ID $k$, we then compute its vector embedding $\mathbf{v}_i \in \mathbb{R}^{1 \times 1 \times d}$ as the mean of the vectors in that sentence,
\begin{equation}
    \mathbf{v}_i = \operatorname{mean}\ (\mathbf{H}[k]_{b_i:e_i}).
\end{equation}
\noindent
Here, $b_i$ and $e_i$ are the beginning and end indices of sentence $i$. We then compute the PSS following Eq.~\eqref{eqa:pss}, and define the redundant sentence set as 
\begin{align}
\begin{aligned}
    \mathcal{P} = \{i: &\lambda_{i,j} > \tau, i\le j\}, \\ \text{where}&\ 
    \lambda_{i,j} = \mathbf{v}_i^\top\mathbf{v}_j.
\end{aligned}
\end{align}
This means that if a pair $(i,j)$ exceeds a predefined threshold $\tau$ (e.g., 0.95), the earlier sentence $i$ is flagged as redundant, while the later one $j$ is retained.
\paragraph{Token Importance Score.}  
Let $\mathbf{Q}\in\mathbb{R}^{bs\times H_q\times \alpha\times d}$ denote the observation window of recent $\alpha$ query tokens \cite{li2024snapkv}, and $\mathbf{K},\mathbf{V}\in\mathbb{R}^{bs\times H_k\times N'\times d}$ denote the current key and value cache, respectively.
Here, $H_q$ and $H_k$ represent the number of query and key-value heads.
Most LRMs use group-query attention, where
for each query head $i\in\{1,\dots,H_q\}$, its associated key-value head is denoted by $g(i)\in\{1,\dots,H_k\}$, where $g(i) = \lfloor\frac{i}{n}\rfloor$. Here, $n$ is an integer that defines the number of query heads over which each K/V is shared.
The attention importance for each query head $i$ in a batch is denoted as $\mathbf{A}^{i} \in \mathbb{R}^{bs\times 1\times \alpha \times N'}$, where
\begin{align} 
    \mathbf{A}^{i} = \text{softmax}(\frac{\mathbf{Q}^{i}\mathbf{K}^{g(i)\top}}{\sqrt{d}} + \mathbf{M}) 
\end{align}
Here, $\mathbf{M}$ denotes the attention mask used for multi-batch decoding.
We then compute the attention importance for a key head $h_k$ as
$\mathbf{I}^{h_k}$ = softmax(maxpool($\mathbf{A}^{h_k\cdot n}$: $\mathbf{A}^{h_k\cdot n+(n-1)}$)). We normalize the attention importance matrix on the observation window of $\alpha$, for each head, to produce the token importance matrix $\mathbf{I}_{\alpha}^{h_k} \in \mathbb{R}^{bs\times N'}$.

\paragraph{Token Redundancy Score.} Inspired by R-KV, we summarize the token redundancy removal method. For each key-value head $h_k$, we define the key state as $\mathbf{K}^{h_k} \in \mathbb{R}^{bs\times 1 \times N' \times d}$, and the token redundancy $\mathbf{R}^{h_k} \in \mathbb{R}^{bs\times N'}$ is calculated as 
\begin{equation}
\begin{aligned}
\mathbf{R}^{h_k} = \operatorname{mean}\Big(\text{softmax}(\bar{\mathbf{K}}^{h_k}\bar{\mathbf{K}}^{h_k}{^\top})\Big), \\ \text{where}\ 
\bar{\mathbf{K}}^{h_k} = \frac{\mathbf{K}^{h_k} \odot \mathbf{M}}{||\mathbf{K}^{h_k} \odot \mathbf{M}||_2 + \epsilon}.
\end{aligned}
\end{equation}
Here, we also consider the attention mask $\mathbf{M}$ to reduce the influence of the padding tokens on the redundancy score.
\paragraph{Sentence Redundancy Driven Cumulative Score.}
We then formulate the overall eviction score for KV compression (Steps 3b in Fig.~\ref{fig:overview}) by combining the token importance score ${I^{h_k}_{\alpha}}$, token redundancy score $R^{h_k}$, and sentence similarity score $\mathbf{\lambda}_{i,j}$:
\begin{equation}
I_{\text{final}} = 
\begin{cases}
\sigma I^{h_k}_{\alpha} - (1 - \sigma) R^{h_k} - \lambda_{i,j}, & \text{if } i \in \mathcal{P},\\[3pt]
\sigma I^{h_k}_{\alpha} - (1 - \sigma) R^{h_k}, & \text{otherwise.}
\end{cases}
    \label{eqa:score}
\end{equation}
$\sigma$ controls the trade-off between token importance and redundancy. As described earlier, $\mathbf{\lambda}_{i,j}$ is the sentence redundancy score associated with sentence pair $(i, j)$, shared over all tokens of the $i^{th}$ sentence.
To meet a specific token budget, we evict tokens in increasing order of final score $I_{\text{final}}$.
Importantly, \textit{since similarity scores for redundant sentences ($\geq$ 0.95) are typically an order of magnitude higher than token-level scores ($\sim 0.1$), Eq. \eqref{eqa:score} ensures that highly redundant sentences are removed before token level eviction}.

\begin{figure}[!t]
    \centering
    \includegraphics[width=.90\linewidth]{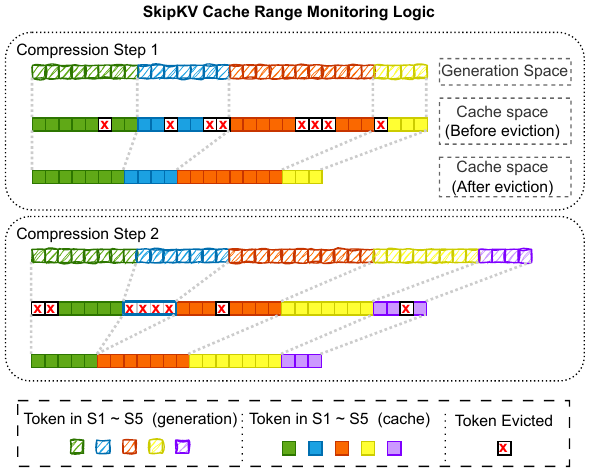}
    \vspace{-2mm}
    \caption{
        Illustration of the cache range monitoring process over two consecutive compression steps.
        Colored blocks represent distinct sentence spans in the generation space and their corresponding regions in the KV cache.
        Gray dashed lines indicate the mapping of sentence range.
    }
    \vspace{-5mm}
    \label{fig:kv_cache_monitor}
\end{figure}

\paragraph{KV Cache Sentence Range Monitoring Logic.}
To ensure that sentences in the redundancy set $\mathcal{P}$ are evicted consistently from the cache space, we introduce a \emph{KV-cache range monitoring mechanism} (Steps 3a and 3c in Fig.~\ref{fig:overview}). 
This mechanism is defined by a mapping function $\Phi$, that converts the $i^{th}$ sentence span in the original generation space (gs) to that in the cache space (cs), $ \Phi $($b^{(gs)}_{i}$, $e^{(gs)}_{i}$) $\rightarrow$ ($b^{(cs)}_{i}$, $e^{(cs)}_{i}$).
To correctly associate sentence-level scores for each sentence in $\mathcal{P}$, we record the token span ids ($b^{(gs)}_{i}$, $e^{(gs)}_{i}$)  and the corresponding PSS in a lookup table $\mathcal{T}$ as
\begin{align}
    \mathcal{T} \gets \{(b^{(gs)}_{i}, e^{(gs)}_{i}) \mapsto \mathbf{\lambda}_{i,j}\}, \quad \forall i \in \mathcal{P}.
    \label{eqa:sentence-score-input}
\end{align}
By applying the mapping function $\Phi$, we obtain the \textbf{cache-aligned lookup table} as
\begin{align}
    \Phi(\mathcal{T}) \rightarrow \{(b^{(cs)}_{i}, e^{(cs)}_{i}) \mapsto \mathbf{\lambda}_{i,j}\}, \quad \forall i \in \mathcal{P}.
    \label{eqa:sentence-score-cache}
\end{align}
The cache-aligned lookup table $\Phi(\mathcal{T})$ is then used in Eq.~\eqref{eqa:score} 
to determine whether $i \in \mathcal{P}$ during eviction.

As illustrated in Fig.~\ref{fig:kv_cache_monitor}, the sentence range monitoring logic maintains a dynamic mapping between the generation space (gs) and the cache space (cs) throughout the compression process. After compression step 1, the retained tokens serve as the initial cs tokens for step 2, with the mapping continuously updated to correctly align gs indices with their corresponding positions in cs.

\emph{Updating Cache Ranges Before Eviction.}
As eviction is performed periodically during decoding, newly generated tokens—often initiating or completing sentences—may be appended to the gs tokens at each eviction step.
Accordingly, their corresponding sentence spans are also added to the cs, following two cases:
(1) initialization of sentence ranges during the first step (Fig.~\ref{fig:kv_cache_monitor}, top), and
(2) appending new ranges in subsequent steps (Fig.~\ref{fig:kv_cache_monitor}, bottom).
In case~(1), the cache ranges are simply initialized to those in the gs:
\begin{align}
    (b^{(cs)}_{i}, e^{(cs)}_{i}) \gets (b^{(gs)}_{i}, e^{(gs)}_{i}), \forall i
\label{eqa:init}
\end{align}

At compression step~$t$, we denote the \textit{generation length} and \textit{cache length before eviction} as $l_t^{(\text{gs})}$ and $l_t^{(\text{cs})}$, while the \textit{post-eviction cache length} is constrained by the budget~$B$.
In case~(2), for each new sentence added during the current compression step, we sequentially update the sentence range in increasing order of sentence index. For the $i^{\text{th}}$ one, it is,
\begin{align}
\begin{split}
&b_{i}^{(\text{cs})} \gets e_{i-1}^{(\text{cs})} + 1, \\
&e_{i}^{(\text{cs})} \gets l_{t}^{(\text{cs})} - \Delta,
\text{where } \Delta = l_{t}^{(\text{gs})} - e_{i}^{(\text{gs})}.
\label{eqa:before}
\end{split}
\end{align}
The end position $e_i^{(\text{cs})}$ is determined by subtracting the offset $\Delta$ from the current cache length $l_t^{(\text{cs})}$. 
Here, $\Delta$ represents the \textit{length of residual tokens} in the gs starting from the $(i+1)^{\text{th}}$ sentence.
This offset is equivalent to the length of the newly appended tokens and can be computed as the difference between the total generation length and the end position of the $i^{\text{th}}$ sentence in the generation space.

\emph{Updating Cache Ranges After Eviction.}
As shown in Fig. \ref{fig:kv_cache_monitor}, sentence spans must be remapped into the new cache coordinate space according to the surviving token indices in each compression step.
Let the set of surviving token indices be
$P = \{p_1, p_2, \dots, p_B\}$ with $0 \leq p_1 < p_2 < \cdots < p_B < l_t^{(cs)}$, with $p_k$ denoting the $k$-th surviving token index. 
The corresponding cache ranges are then updated as
\begin{align}
\begin{split}
&b_i^{(\text{cs})} \leftarrow \min \{ k \mid p_k \geq b_i^{(\text{cs})} \}, \\
&e_i^{(\text{cs})} \leftarrow \max \{ k \mid p_k \leq e_i^{(\text{cs})} \}, \forall i
\label{eqa:after}
\end{split}
\end{align}
Here, $b_i^{(\text{cs})}$ is reassigned to the earliest surviving index not earlier than the original start, and $e_i^{(\text{cs})}$ to the latest surviving index not later than the original end.
If no tokens in the compressed cache space survive, the corresponding sentence range is discarded from the post-eviction cache space. 

\begin{figure}[t]
    \centering
    \includegraphics[width=\linewidth]{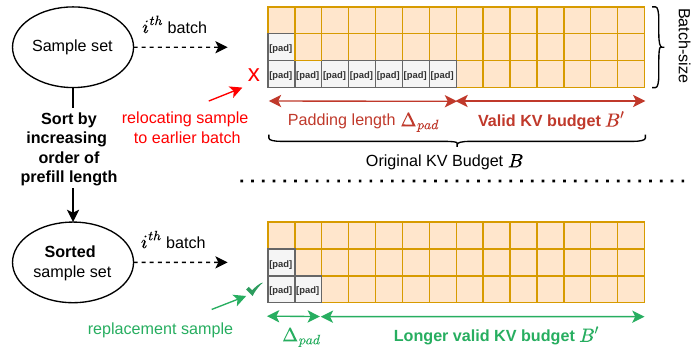}
    \vspace{-5mm}
    \caption{Batch grouping increases valid KV budget for high-performance multi-batch decoding. Yellow blocks: valid tokens in the KV cache, Gray blocks: padding tokens.}
    \vspace{-5mm}
    \label{fig:grouping}
\end{figure}

\begin{figure*}[t]
    \centering
    \includegraphics[width=.9\linewidth]{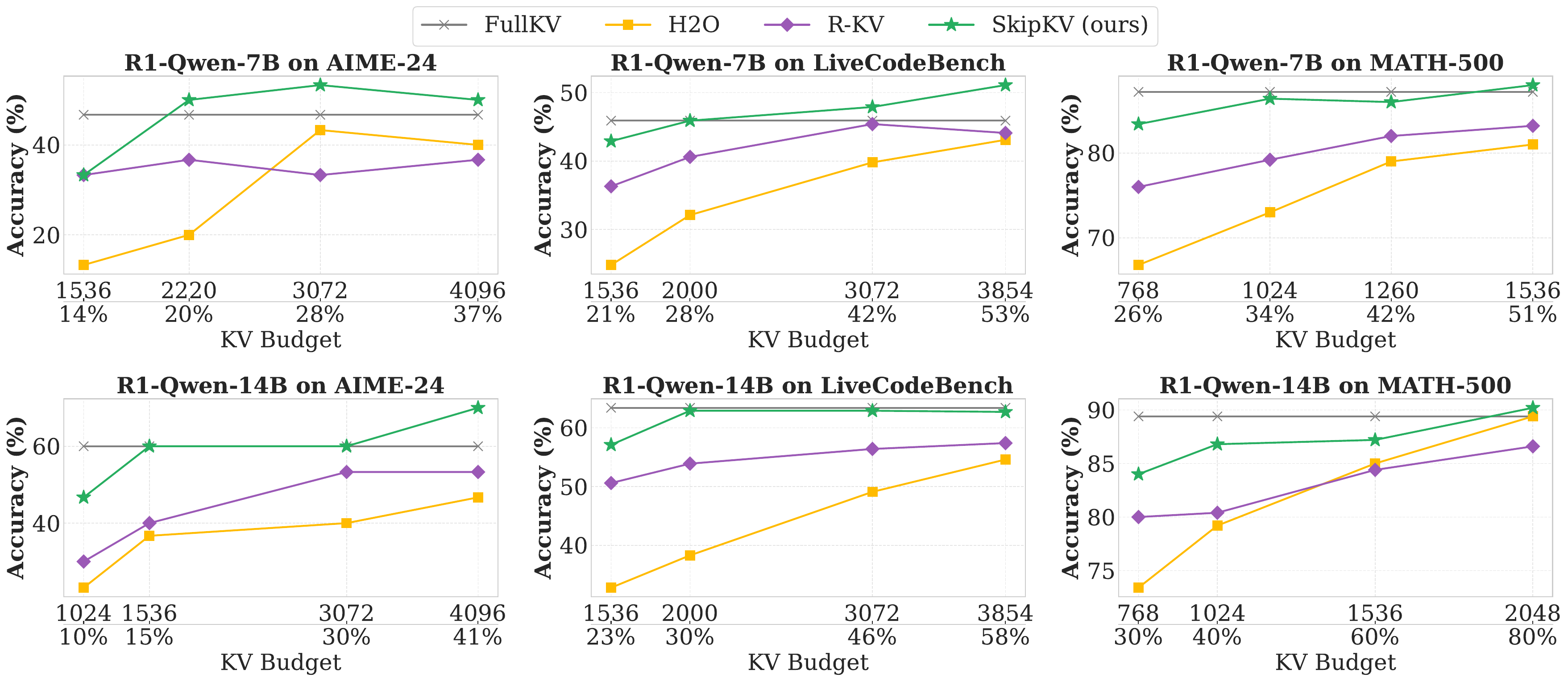}
    \vspace{-3mm}
    \caption{Accuracy comparison under different KV-cache budgets for SkipKV, H2O, R-KV, and FullKV across three reasoning benchmarks and R1-Qwen-7B and 14B models. SkipKV consistently achieves higher accuracy under tighter KV budgets, maintaining full accuracy even at only 15 \% KV budget on AIME-24. Results are reported as pass@1.}
    \vspace{-5mm}
    \label{fig:main_acc}
\end{figure*}

\subsection{Skipping KV Generation with Adaptive Steering}
\label{sec:adaptive-steering}
In addition to removing redundant thoughts after their generation, we further propose to \emph{skip unnecessary thoughts before generation} through an adaptive steering mechanism. 
For this, we add a steering vector to the hidden state of certain LRM layer(s), enforcing the model to be more precise. 
However, unlike earlier works on steering \cite{chen2025seal, azizi2025activation} that use a fixed strength ($\alpha_0$) for the steering vector throughout generation, we propose a strength adaptation based on the number of non-execution thoughts. The details are presented below.

Inspired by SEAL \cite{chen2025seal} we first construct the steering vector using 1000 samples from the MATH training set \cite{hendrycks2measuring}, aiming to shift the latent representations towards execution-style reasoning.
Such reasoning-oriented steering directions exhibit strong transferability across tasks, as shown in prior work~\cite{chen2025seal, azizi2025activation} and our experiments.
Importantly, our objective is not to eliminate non-execution thoughts. While execution thoughts correspond to explicit computation, non-execution thoughts can also be beneficial. We therefore treat this distinction as \emph{statistical rather than causal}, and softly bias the model toward patterns more frequently associated with successful reasoning.

The steering vector is computed as the mean latent representation difference between execution and non-execution thoughts:
    $\mathbf{v} = \overline{\mathbf{H}_E} - \overline{\mathbf{H}_O},$
where $\overline{\mathbf{H}_E}$ and $\overline{\mathbf{H}_O}$ denote the average hidden states of execution and non-execution thoughts, respectively.
During generation, at each reasoning newline delimiter $y_t \in \mathcal{D}$, we inject the steering vector into the hidden representation of the selected layer $k$, where $\mathcal{D}$ denotes the predefined delimiter set (see Appendix~\ref{sec:appendix-alg}), i.e.,
    $\mathbf{H}_k \leftarrow \mathbf{H}_k + \alpha_t \cdot \mathbf{v}.$
Here, $\alpha_t$ denotes the steering strength controlling the degree of latent adjustment toward execution-oriented directions.

We dynamically adjust the steering strength $\alpha_t$ based on the \textit{model’s observed reasoning behavior}. Specifically, as illustrated in Fig.~\ref{fig:overview} (left), we maintain a running counter of non-execution thoughts, denoted as $N_o$, and update the steering strength as
    $\alpha_t \leftarrow \alpha_0 + \gamma \cdot N_o,$
where $\alpha_0$ is the initial steering strength and $\gamma$ is a predefined increment factor controlling the degree of steering aggressiveness.
In this way, the steering strength for each sample is adaptively adjusted throughout the reasoning process, enabling the model to shorten generation length by selectively suppressing unnecessary thoughts.

\subsection{Batch Grouping for Multi-batch Decoding}
\label{sec:batch-grouping}
For each sample in a batch the \emph{valid} KV budget is given by, $B' = B - \Delta_{pad}$, with $\Delta_{pad}$ being the sample's padding token count.  For any sample, $\Delta_{pad}$ is given by, 
\begin{align}
\vspace{-5mm}
\Delta_{pad} = N^{max}_p - N_p,
\vspace{-5mm}
\label{eqa:batch_padding}
\end{align}
\noindent
where $N_p$ and $N_p^{\max}$ denote the prefill length of the current sample and the maximum prefill length within a batch, respectively. 
When prefill lengths vary significantly across samples, the resulting padding overhead $\Delta_{\text{pad}}$ can be substantial, reducing the effective KV budget. 

To mitigate this inefficiency, we propose a batch grouping strategy that minimizes wasteful padding.
Specifically, we first \textit{sort} all samples in increasing order of their prefill lengths, and then \textit{group} them into batches of size $bs$ based on this ordering.
As shown in Eq.~\eqref{eqa:batch_padding}, reducing the gap between $N_p^{\max}$ and $N_p$ lowers the padding overhead, thereby increasing the effective budget $B'$. Fig.~\ref{fig:grouping} illustrates an example of this sample rearrangement. 
In §\ref{sec:ablation} we empirically validate the effectiveness of increased $B'$  in improving LRM accuracy for multi-batch decoding.

\section{Experiments}
In this section, we first compare the accuracy and generation length of SkipKV on complex reasoning tasks against recent token-eviction methods across multiple reasoning models and KV-budget ratios.
We then compare SkipKV with other efficient reasoning baselines, followed by an evaluation of reasoning throughput and ablation studies on each component of SkipKV, including its impact on the effective KV budget under batch grouping.
Additional experimental details and evaluation results are provided in the Appendix.

\subsection{Experimental Setup.}
\paragraph{Models and Datasets.}
We conduct evaluations on DeepSeek-R1-Distill-Qwen-7B, DeepSeek-R1-Distill-Qwen-14B, and DeepSeek-R1-Distill-Llama-8B \cite{guo2025deepseek}. 
Our experiments cover both mathematical reasoning benchmarks—MATH-500~\cite{lightman2023let}, AIME~\cite{MAA_AIME2024}, and GSM8K~\cite{cobbe2021training}—and code generation using LiveCodeBench~\cite{jain2024livecodebench}. 
We constrain the maximum generation length to 8,192 tokens for GSM8K and MATH-500, 10,000 tokens for LiveCodeBench, and 16,384 tokens for AIME-24.
Following SEAL~\cite{chen2025seal}, we derive the steering vector using 1,000 samples from the training set of the Math dataset \cite{hendrycks2measuring}.
Unless otherwise stated, the evaluation batch size is set to 10 for R1-Qwen-7B and R1-Llama-8B, while for R1-Qwen-14B, it is set to the maximum that can fit into GPU memory.
All experiments are conducted on a single NVIDIA A100 (40GB) GPU.




\begin{figure*}[t]
    \centering
    \includegraphics[width=.9\linewidth]{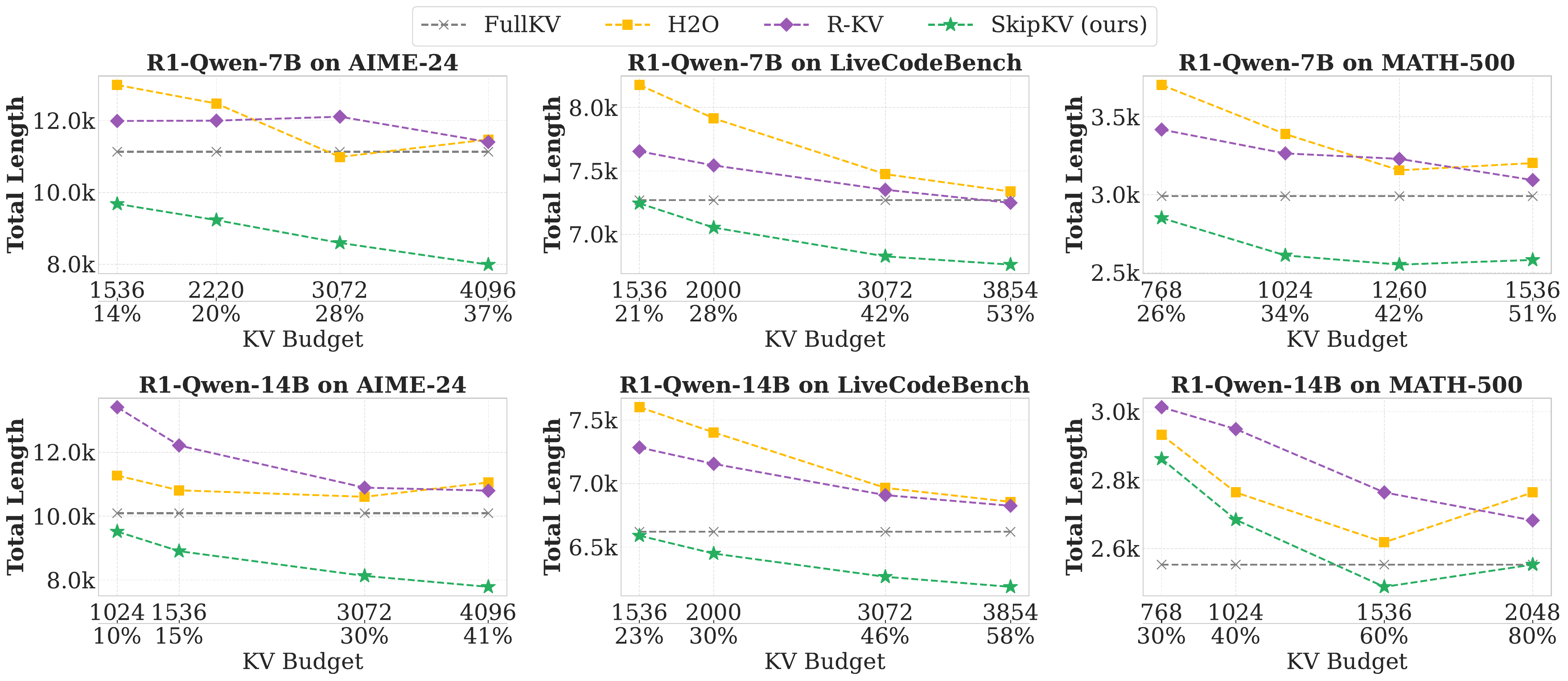}
    \vspace{-3mm}
    \caption{Total generation tokens on SkipKV under different KV budget with H2O, R-KV, and FullKV across three datasets and two models. SkipKV consistently generates fewer tokens and could achieve up to 30\% fewer generation length compared with FullKV.}
    \vspace{-5mm}
    \label{fig:main_token}
\end{figure*}

\subsection{Main Results}
\paragraph{Comparison of Accuracy with Eviction Methods.}
We compare the reasoning accuracy of SkipKV with prior KV cache eviction methods, including H2O, R-KV, and the FullKV baseline across multiple reasoning datasets and model scales, as shown in Fig.~\ref{fig:main_acc} and Fig.~\ref{fig:main_llama} in Appendix~\ref{sec:appendix-main}.
Following R-KV, we define the KV cache budget ratio as the ratio of the allocated KV cache budget to the average generation token length under FullKV for each model–dataset pair.
Different from prior token-level KV eviction methods, which suffer from severe accuracy degradation during multi-batch decoding, our SkipKV approach consistently maintains high accuracy with significantly lower KV memory across all reasoning tasks and models.
On challenging reasoning tasks such as AIME-24, SkipKV matches the FullKV accuracy using $6.7 \times$ lower KV cache memory on R1-Qwen-14B, while maintaining more than $4 \times$ lower KV memory usage across all models.
In addition, SkipKV could achieve better performance compared with FullKV on a limited KV budget.
For example, SkipKV improves accuracy by 5.2\% on LiveCodeBench with $2\times$ less KV memory evaluated on R1-Qwen-7B, and by 10\% on AIME-24 while using $2.5\times$ lower KV memory evaluated on R1-Qwen-14B.


\paragraph{Comparison of Token Length with Eviction Methods.}
Besides achieving higher reasoning accuracy, SkipKV also provides substantial generation efficiency gains by producing fewer tokens.
As shown in Fig. \ref{fig:main_token}, prior token-level eviction methods consistently generate more tokens than FullKV across all models and tasks, whereas SkipKV reduces the total token length by up to 28\% compared with FullKV.
Compared with R-KV, SkipKV generates up to 32\%, 39\%, and 48\% fewer tokens on R1-Qwen-7B, R1-Qwen-14B, and R1-Llama-8B, respectively—translating to a $1.5-2\times$ reduction in generation latency.
Additionally, on the MATH-500 dataset, SkipKV produces approximately 15\% 
fewer tokens while using $3\times$ less KV memory than FullKV on R1-Qwen-7B resulting in $3.5\times$ overall memory-latency benefits. Similarly, on LiveCodeBench, SkipKV consistently achieves around 10\% shorter generation lengths across all KV budget settings while maintaining comparable accuracy to that with FullKV.

\begin{figure*}[htbp]
    \centering
    \includegraphics[width=\linewidth]{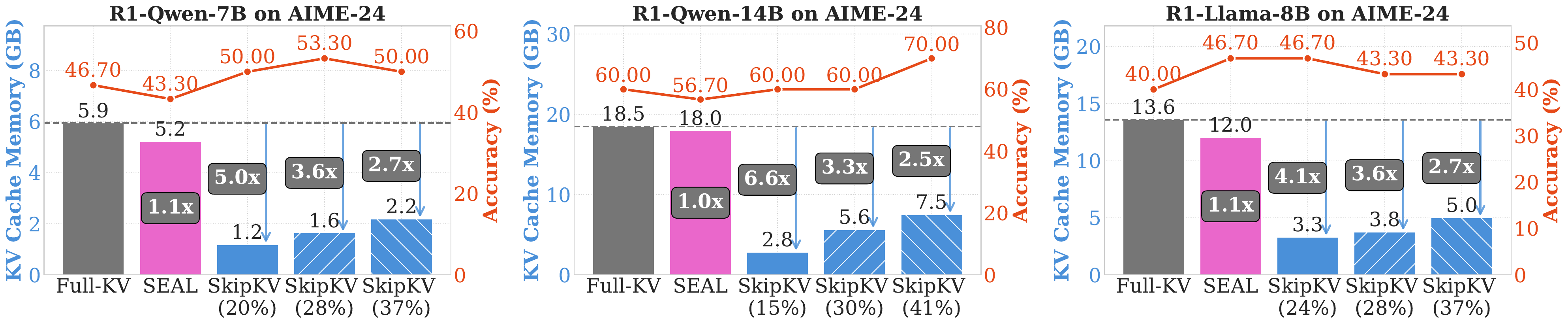}
    \vspace{-7mm}
    \caption{KV-cache memory consumption and reasoning accuracy of SkipKV under different KV budgets, compared with SEAL and Full-KV baselines on AIME-24 across multiple reasoning models.
    \textbf{Left}: R1-Qwen-7B; \textbf{Center}: R1-Qwen-14B; \textbf{Right}: R1-Llama-8B.}
    \vspace{-5mm}
\label{fig:kvmemory}
\end{figure*}

\paragraph{Comparison with Other Efficient Reasoning Methods.}
We further compare SkipKV with SEAL that focus on reducing generation length. 
Fig.~\ref{fig:kvmemory} presents the comparison of KV-cache memory consumption and reasoning accuracy among SkipKV, SEAL, and FullKV on the AIME-24 dataset and across three reasoning models.
We set the steering factor of SEAL to $1$, following its original configuration.
Because SEAL does not explicitly compress the KV cache, its KV memory footprint is approximated using the average number of active tokens per sequence, which is proportional to the mean generation length.
While SEAL successfully shortens reasoning traces and maintains accuracy, its modest reduction in output token length (about 10\%) translates to only limited KV memory savings.
In contrast, SkipKV jointly skips KV generation and storage, achieving up to $13.3\%$ accuracy improvement and $6.6 \times$ KV memory reduction  when compared with SEAL.
These results highlight that SkipKV effectively balances memory compression and reasoning fidelity across different reasoning models.

\begin{table}[t]
\centering
\caption{Comparison of total latency and throughput under different batch sizes on GSM8K. Evaluated on one A100-40GB.}
\vspace{2mm}
\resizebox{0.45\textwidth}{!}{
\begin{tabular}{l|c|c|c|c|c|c}
\toprule
\multicolumn{1}{c|}{\textbf{Metric}} & \textbf{Batch-size} &\textbf{10}  & \textbf{50} & \textbf{100}  & \textbf{120} & \textbf{140} \\ 
\midrule
\multirow{4}{*}{\shortstack{\textbf{Total Latency}\\(min) $\downarrow$}} 
 & FullKV & 324 & 500 & OOM & OOM & OOM \\ \cline{2-7}
 & SEAL   & 178 & 192 & OOM & OOM & OOM\\ 
 & R-KV   & 227 & 96  & {66} & 73 & 70  \\ 
\rowcolor{blue!5}
 & \textbf{SkipKV (ours)} & \textbf{136} & \textbf{58}  & \textbf{52} & \textbf{66} & \textbf{68} \\ 
\midrule
\multirow{4}{*}{\shortstack{\textbf{Throughput}\\(samples/min) $\uparrow$}}
 & FullKV & 4.07 & 2.64 & OOM & OOM & OOM \\ \cline{2-7}
 & SEAL   & 7.41 & 6.87 & OOM & OOM & OOM \\ 
 & R-KV   & 5.81 & 13.7 & {20.0} & 18.1 & 18.8\\ 
\rowcolor{blue!5}
 & \textbf{SkipKV (ours)}  & \textbf{9.70} & \textbf{22.7} & \textbf{25.4} & \textbf{20.0} & \textbf{19.4}\\ 
\bottomrule
\end{tabular}
}
\label{table:throughput}
\vspace{-5mm}
\end{table}

\begin{table*}[htbp]
\centering
\caption{
Ablation of SkipKV evaluated on AIME24 using R1-Qwen-7B. 
Progressive inclusion of Sentence Scoring, Adaptive Steering, and Batch Grouping 
improves accuracy and reduces total token length compared with the R-KV baseline.
}
\vspace{2mm}
\scriptsize
\setlength{\tabcolsep}{5pt}
\resizebox{.9\textwidth}{!}{
\begin{tabular}{l|ccc|ccc}
\toprule
 & \multicolumn{3}{c|}{\textbf{Accuracy (\%)} $\uparrow$} & \multicolumn{3}{c}{\textbf{Total Token Length} $\downarrow$} \\ \hline
\textbf{KV Budget} & 2220 (20\%) & 3072 (27\%) & 4096 (37\%) 
& 2220 (20\%) & 3072 (27\%) & 4096 (37\%) \\ \hline
FullKV & \multicolumn{3}{c|}{46.7} & \multicolumn{3}{c|}{11132} \\ \hline
R-KV   & 36.7 & 33.3 & 36.7 & 12000 & 12109 & 11403 \\
+ Sentence Scoring & 40.0 (+3.3) & 36.7 (+3.4) & 40.0 (+3.3) & 11332 (–6\%) & 11342 (–6\%) & 11819 (+4\%) \\
+ Adaptive Steering & 40.0 (+3.3) & 43.3 (+10.0) & 40.0 (+3.3) & 8860 (–26\%) & 10101 (–17\%) & 10041 (–12\%) \\
\rowcolor{blue!7}
+ Batch Grouping (SkipKV) & \textbf{50.0 (+13.3)} & \textbf{53.3 (+20.0)} & \textbf{50.0 (+13.3)} 
                          & \textbf{9228 (–23\%)} & \textbf{8593 (–29\%)} & \textbf{7988 (–30\%)} \\
\bottomrule
\end{tabular}}
\label{table:component}
\end{table*}

\begin{figure}[t]
    \centering
    \includegraphics[width=\linewidth]{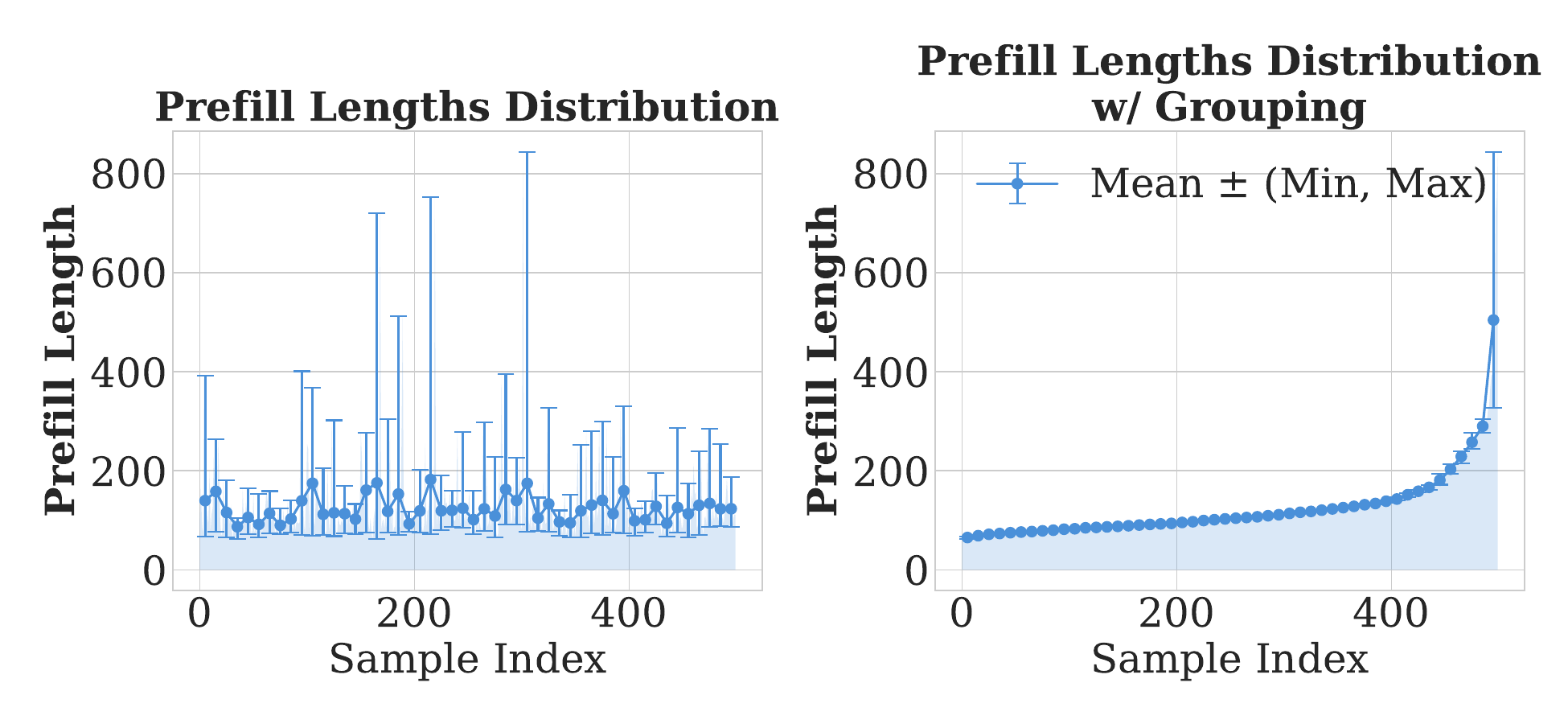}
    \vspace{-7mm}
    \caption{Visualization of the prefill token length distribution of MATH-500 and the min-max range of each 10 samples before (left) and after (right) using batch grouping.}
    \vspace{-5mm}
    \label{fig:multi-batch}
\end{figure}

\paragraph{Throughput Analysis.}
We evaluate the end-to-end throughput of different methods on the GSM8K dataset using R1-Qwen-7B with a maximum generation length of 8,192 tokens on a single A100-40GB GPU.
The KV-cache budget is set to 512, under which both SkipKV and R-KV achieve comparable accuracy to FullKV.
We measure the total latency and throughput (samples processed per minute) across various batch sizes, with FlashAttention-2 enabled for all methods. 
As shown in Table~\ref{table:throughput}, we define the transition point as the batch size at which each method achieves its peak throughput—beyond which performance becomes memory-bounded.
We observe that FullKV and SEAL support significantly smaller batch sizes than the eviction-based methods, resulting in earlier transition points and limited throughput scalability.
In contrast, R-KV and SkipKV effectively alleviate memory pressure via their fixed-size KV caches, enabling up to $2.8\times$ larger batch sizes and substantially higher throughput.
Overall, R-KV achieves up to a $7.6\times$ speedup over FullKV, while SkipKV further accelerates generation by $9.6\times$ compared with FullKV.
Additionally, at the same batch size, SkipKV outperforms R-KV by up to $1.7\times$, benefiting from its shorter generation length. 

\subsection{Ablation Studies}
\label{sec:ablation}
\paragraph{SkipKV Components.}

From Table \ref{table:component}, we observe that progressively integrating the three SkipKV components—Sentence Scoring ($\S$~\ref{sec:sentence-scoring}), Adaptive Steering ($\S$~\ref{sec:adaptive-steering}), and Batch Grouping ($\S$~\ref{sec:batch-grouping})—yields consistent improvements in both reasoning efficiency and accuracy on the AIME24 benchmark.
Sentence Scoring provides moderate gains by removing redundant sentences, slightly improving accuracy and shortening the generated sequence length.
Including Adaptive Steering further enhances efficiency by dynamically skipping non-execution thoughts, leading to a substantial reduction in total token length without sacrificing accuracy.
Finally, combining Batch Grouping with the previous components forms the complete SkipKV configuration, which achieves the strongest overall results, improving accuracy by up to +20 \% and reducing total token length by as much as 30 \% across different KV budgets compared with the R-KV baseline.
These results demonstrate that SkipKV’s hierarchical design—progressing from local semantic filtering to global decoding coordination—effectively balances reasoning quality and efficiency in large-scale mathematical reasoning tasks.

\begin{table}[t]
\centering
\caption{
Effect of Batch Grouping evaluated on MATH-500 using R1-Qwen-7B with SkipKV.
}
\vspace{2mm}
\scriptsize
\setlength{\tabcolsep}{5pt}
\resizebox{.48\textwidth}{!}{
\begin{tabular}{l|cccc}
\toprule
\textbf{KV Budget} & 768 (26\%) & 1024 (34\%) & 1260 (42\%) & 1536 (51\%) \\
\midrule
\textbf{Avg. Valid Budget} $\uparrow$ & & & & \\
\quad w/o Batch Grouping & 630 (21\%) & 886 (30\%) & 1122 (38\%) & 1398 (47\%) \\
\rowcolor{blue!7}
\quad w/ Batch Grouping  & \textbf{759 (25\%)} & \textbf{1015 (34\%)} & \textbf{1251 (42\%)} & \textbf{1527 (51\%)} \\
\midrule
\textbf{Accuracy (\%)} $\uparrow$ & & & & \\
\quad w/o Batch Grouping & 77.8 & 85.2 & 85.8 & 86.0 \\
\rowcolor{blue!7}
\quad w/ Batch Grouping  & \textbf{83.4 (+5.6)} & \textbf{86.4 (+1.2)} & \textbf{86.0 (+0.2)} & \textbf{88.0 (+2.0)} \\
\bottomrule
\end{tabular}}
\label{table:grouping}
\vspace{-6mm}
\end{table}

\paragraph{Impact of Batch Grouping on Valid KV Budget.}
To evaluate the effectiveness of our batch grouping technique in multi-batch decoding, 
we visualize the prefill token length distribution of the MATH-500 dataset before and after applying batch grouping, 
and analyze its impact on the valid KV budget and overall accuracy.
As shown in Fig.~\ref{fig:multi-batch}, the prefill token lengths vary significantly across the dataset, 
resulting in large intra-batch variation. 
However, after sorting and grouping, the prefill lengths increase smoothly from the first to the last sample, 
substantially reducing the variation within each batch.
This reduction in variation decreases the amount of padding required, 
allowing the average valid KV budget to approach the nominal KV budget and thereby reducing the memory overhead caused by padding tokens.
As summarized in Table~\ref{table:grouping}, batch grouping allocates nearly the entire KV budget to valid tokens, 
effectively preserving accuracy under lower KV budgets.

\section{Conclusions}
We introduced SkipKV, a sentence-oriented KV compression framework designed to enhance reasoning efficiency while maintaining accuracy. It selectively skips KV generation and storage to yield lower memory footprint.
Motivated by empirical observations on sentence-level structures in LRM outputs, SkipKV introduces a sentence-primary KV eviction policy and a sentence-type adaptive steering vector for more coherent and efficient generation.
To further yield robust accuracy at multi-batch decoding with SkipKV, we presented a batch grouping strategy to improve the effective KV budget allocation.
Compared to the SoTA alternative of R-KV, SkipKV yields up to {$\mathbf{26.7}\%$} accuracy improvement at similar compression budget while generating up to $\mathbf{1.6}\times$ fewer tokens. Additionally, in multi-batch settings SkipKV yields a throughput improvement of up to $\mathbf{9.6}\times$ compared to the baseline FullKV LRM, at similar accuracy.

\section{Discussions}
Batch grouping method (§\ref{sec:batch-grouping}) in SkipKV relies on request length characteristics and may be less effective under highly variable or extreme long-context scenarios. In such cases, residual padding overhead can still limit efficiency gains. While dynamic grouping strategies that adapt to prompt-length variability could help mitigate this issue, we leave a systematic exploration of such approaches to future work.

\section{Acknowledgments}
Initial ideation and majority of the work was done during Jiayi's internship at Intel. 
This work is co-funded by Intel Strategic Research Sectors (SRS) - Systems Integration SRS \& Devices SRS.
Additionally, we would like to acknowledge Akshat Ramachandran from Georgia Institute of Technology, Stefano Pellerano, Hai Li, and Arnab Raha from Intel for their insightful conversations and feedback throughout the duration of this project. 

\bibliography{ref}
\bibliographystyle{mlsys2026}

\clearpage
\appendix
\section{Appendix}
\begin{algorithm}[!t]
\caption{Skipping KV Cache Storage}
\label{alg:range_monitor}
\begin{algorithmic}[1]
\REQUIRE KV cache length before eviction $l_t^{(\text{cs})}$, KV cache budget $B$, number of sentences $n$, sentence ranges in generation and cache spaces $\mathcal{M}^{(\text{gs})}$, $\mathcal{M}^{(\text{cs})}$, token span-PSS score lookup table $\mathcal{T}$.
\ENSURE 
Compressed cache $\mathbf{K}_{\text{cache}}, \mathbf{V}_{\text{cache}}$, 
updated cache ranges $\mathcal{M}^{(\text{cs})}$.
\STATE
\STATE \textcolor{gray}{\textit{\# Pre-eviction update}}
\STATE \textcolor{gray}{\textit{\# Case (1): Initialize empty cache ranges.}}
\IF{$\mathcal{M}^{(\text{cs})}$ is empty}
    \FOR{each sentence range $\mathcal{M}_i^{(\text{gs})}$ where $e_i^{(\text{gs})}$ $\le l_t^{(\text{cs})}$}
        \STATE Initialize $\mathcal{M}_i^{(\text{cs})}$ using Eq.~\eqref{eqa:init}.
    \ENDFOR
\STATE \textcolor{gray}{\textit{\# Case (2): Append new sentence ranges.}}
\ELSE 
    \FOR{each sentence range $\mathcal{M}_i^{(\text{gs})}$ where $i \ge n$}
        \STATE Update $\mathcal{M}_i^{(\text{cs})}$ using Eq.~\eqref{eqa:before}.
        \STATE \mbox{$\mathcal{M}^{(\text{cs})} \leftarrow \mathcal{M}^{(\text{cs})} \cup \{\mathcal{M}_i^{(\text{cs})}\}$, \textbf{if} $e_i^{(\text{cs})} \le l_t^{(\text{cs})}$.}

    \ENDFOR
\ENDIF
\STATE
\STATE \textcolor{gray}{\textit{\# KV eviction}}
\STATE Map redundancy scores to cache ranges $\Phi(\mathcal{T})$ via Eq.~\eqref{eqa:sentence-score-cache}; evict redundant tokens and update $\mathbf{K}_{\text{cache}}, \mathbf{V}_{\text{cache}}$ with fixed budget $B$ via Eq.~\eqref{eqa:score}.
\STATE 
\STATE \textcolor{gray}{\textit{\# Post-eviction update}}
\STATE Update $\mathcal{M}_i^{(\text{cs})}$ using Eq.~\eqref{eqa:after} $, \forall \mathcal{M}_i^{(\text{cs})} \in \mathcal{M}^{(\text{cs})}$.
\STATE Update the sentence count $n \leftarrow \text{len}(\mathcal{M}^{(\text{gs})})$.
\end{algorithmic}
\end{algorithm}

\begin{algorithm}[!t]
\caption{SkipKV Algorithm}
\label{alg:skipkv}
\begin{algorithmic}[1]
\REQUIRE 
Large Reasoning Model $f(\theta, \cdot)$, 
input content $X$, 
KV cache budget $B$, 
compression step interval $\Delta_t$, 
newline delimiter set $\mathcal{D}$,
non-executable keyword set $\mathcal{N}$,
steering layer index $L_s$,
steering increment factor $\gamma$,
initial steering strength $\alpha_0$,
steering vector $\mathbf{v}$,
maximum generation length $N$
\ENSURE Generated text $Y$.
\WHILE{$t < N$}
\STATE \textcolor{gray}{\textit{\#  $1\sim 2$. Record labeled generated sentence ranges}}
\STATE $i \leftarrow 0, b_i^{(gs)} \leftarrow 0$
\FOR{$x_t$ in X}
\IF{$x_t \in \mathcal{D}$}
    \STATE $\mathcal{M}_i^{(gs)} \leftarrow (b_i^{(gs)}, x_t)$ , $b_{i+1}^{(gs)} \leftarrow x_t + 1, i \leftarrow i + 1$ 
    \IF{$(\forall x_t \in \mathcal{M}_i^{(gs)})\in \mathcal{N}$}
        \STATE Label $\mathcal{M}_i^{(gs)}$ with 'Non-exe' 
    \ENDIF
\ENDIF
\ENDFOR
\STATE 

\STATE \textcolor{gray}{\textit{\#  3. Skip KV Storage}}
\STATE $\mathbf{K}_{\text{cache}}, \mathbf{V}_{\text{cache}}, y \leftarrow f(\theta, X)$
\FOR{$k$ in decoder layers $L$}
\IF{$t \bmod \Delta_t == 0$}
    \STATE Update $\mathbf{K}_{\text{cache}}, \mathbf{V}_{\text{cache}}$, and $\mathcal{M}^{(\text{cs})}$ using Alg. \ref{alg:range_monitor} 
\ENDIF
    \STATE \textcolor{gray}{\textit{\# Skip KV Generation}}
    \STATE $\mathbf{H}_k \leftarrow \mathbf{H}_k + \alpha_t \cdot \mathbf{v}$, \textbf{if} $k == L_s \land x_t \in \mathcal{D}$
\ENDFOR

\STATE 
\STATE \textcolor{gray}{\textit{\#  $4\sim 5$. Compute PSS}}
    \STATE Compute redundancy scores for sentences in $\mathcal{M}^{(\text{gs})}$ 
    and record redundant set $\mathcal{T}$ using Eq.~\eqref{eqa:sentence-score-input}.

\STATE 
\STATE \textcolor{gray}{\textit{\# Update Steering Strength}}
\STATE $N_o \leftarrow \text{count}(\mathcal{M}_i^{(\text{gs})}\text{.key()} == \text{'Non-exe'}, \forall i)$
\STATE $\alpha_t \leftarrow \alpha_0 + N_o \cdot \gamma$

\STATE 
\STATE \textcolor{gray}{\textit{\# Auto-regressive update}}
\STATE $y \!\rightarrow\! Y;\quad y \!\rightarrow\! X;\quad t \leftarrow t + 1$
\ENDWHILE

\end{algorithmic}
\end{algorithm}

\subsection{SkipKV: Algorithm}
\label{sec:appendix-alg}
We provide the pseudo-code of the SkipKV storage-skipping mechanism in Algorithm~\ref{alg:range_monitor}. 
Suppose there are totally $n$ sentences, we define the sets of sentence spans in the generation space and their corresponding ranges in the KV cache as
\begin{equation}
\mathcal{M}^{(\text{gs})} = \{(b_i^{(\text{gs})}, e_i^{(\text{gs})})\}_{i=1}^{n},
\quad
\mathcal{M}^{(\text{cs})} = \{(b_i^{(\text{cs})}, e_i^{(\text{cs})})\}_{i=1}^{n},
\end{equation}
where the $i^{th}$ sentence span in the original generation space (gs) should be mapped to that in the cache space (cs) as $ \Phi $($b^{(gs)}_{i}$, $e^{(gs)}_{i}$) $\rightarrow$ ($b^{(cs)}_{i}$, $e^{(cs)}_{i}$) using the cache range monitoring logic mentioned in §\ref{sec:sentence-scoring}.
At each eviction step, we obtain the compressed $\mathbf{K}_{\text{cache}}, \mathbf{V}_{\text{cache}}$, and the updated sentence ranges in cache space $\mathcal{M}^{(\text{cs})}$ with Algorithm \ref{alg:range_monitor}.

\begin{figure*}[htbp]
    \centering
    \includegraphics[width=\linewidth]{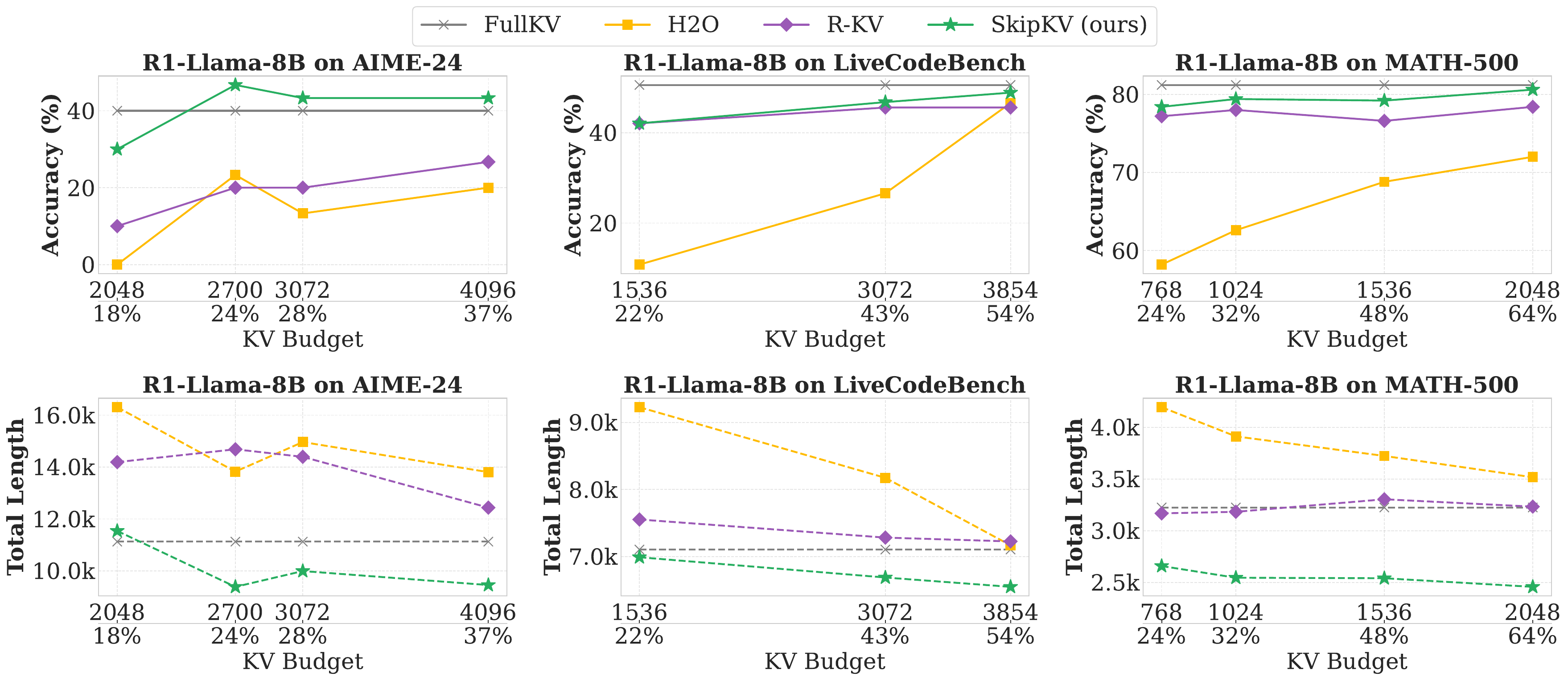}
    \vspace{-7mm}
    \caption{Comparison of accuracy and total generation token length on SkipKV under different KV budget with H2O, R-KV, and FullKV across three datasets and R1-Llama-8B model.}
    \label{fig:main_llama}
    \vspace{-3mm}
\end{figure*}

The complete SkipKV procedure including both KV storage skipping and KV generation skipping is detailed in Algorithm~\ref{alg:skipkv}.
For clarity, we define the \textbf{newline delimiter set}, which specifies where sentence ranges are monitored and steering vectors are applied, as follows:
$$\mathcal{D} = \{"\texttt{\textbackslash n}","\ \texttt{.\textbackslash n}",\ "\texttt{)\textbackslash n}","\ \texttt{\textbackslash n\textbackslash n}",\ "\texttt{.\textbackslash n\textbackslash n}","\ \texttt{)\textbackslash n\textbackslash n}"\}.$$
To improve robustness, we merge consecutive delimiter tokens to avoid over-fragmentation during sentence segmentation. Additionally, for mixed contexts, the delimiter set can be empirically customized based on the generation domain. In our experiments on LiveCodeBench, we find that the proposed delimiter set remains effective. Importantly, we exclude syntax-critical delimiters (e.g., \texttt{:\textbackslash n}) that commonly appear in control-flow constructs such as \texttt{if}-\texttt{else} or \texttt{for} loops, as splitting on these could fragment semantically coherent code blocks and increase the risk of premature eviction.

The \textbf{non-executable keyword set}, which determines where the steering strength is increased, is defined as:
$$\mathcal{N} = \{\texttt{"Wait"},\ \texttt{"Alternatively"},\ \texttt{"again"}\}.$$
These sets are initialized once before decoding and referenced to record the labeled input sentence ranges.

\subsection{Experimental Setup}
\label{sec:appendix-setup}
\paragraph{Hyper-parameters.}
Following R-KV, we compress and update KV cache every 128 decoding steps and set the attention-redundancy score trade-off factor to $\sigma=0.1$.
The similarity threshold $\tau$ in the sentence-scoring metric is selected within the range of $0.95$ to $0.99$. 
The steering strength $\alpha$ is initialized as $1$ or $1.25$, and the steering strength increment factor $\gamma$ is set to $0.02$. 
We insert the steering vector in the 20-th layer of R1-Qwen-7B and R1-Llama-8B, and in the 35-th layer of R1-Qwen-14B.

\paragraph{Baselines.}
We compare our method against eviction-based KV compression methods H2O \cite{zhang2023h2o}, R-KV~\cite{cai2025r}, and the steering-based efficient reasoning method SEAL~\cite{chen2025seal}. 
FullKV is included as a reference method that keeps the full KV cache, providing the gold standard for decoding accuracy.


\begin{figure*}[htbp]
    \centering
    \includegraphics[width=.9\linewidth]{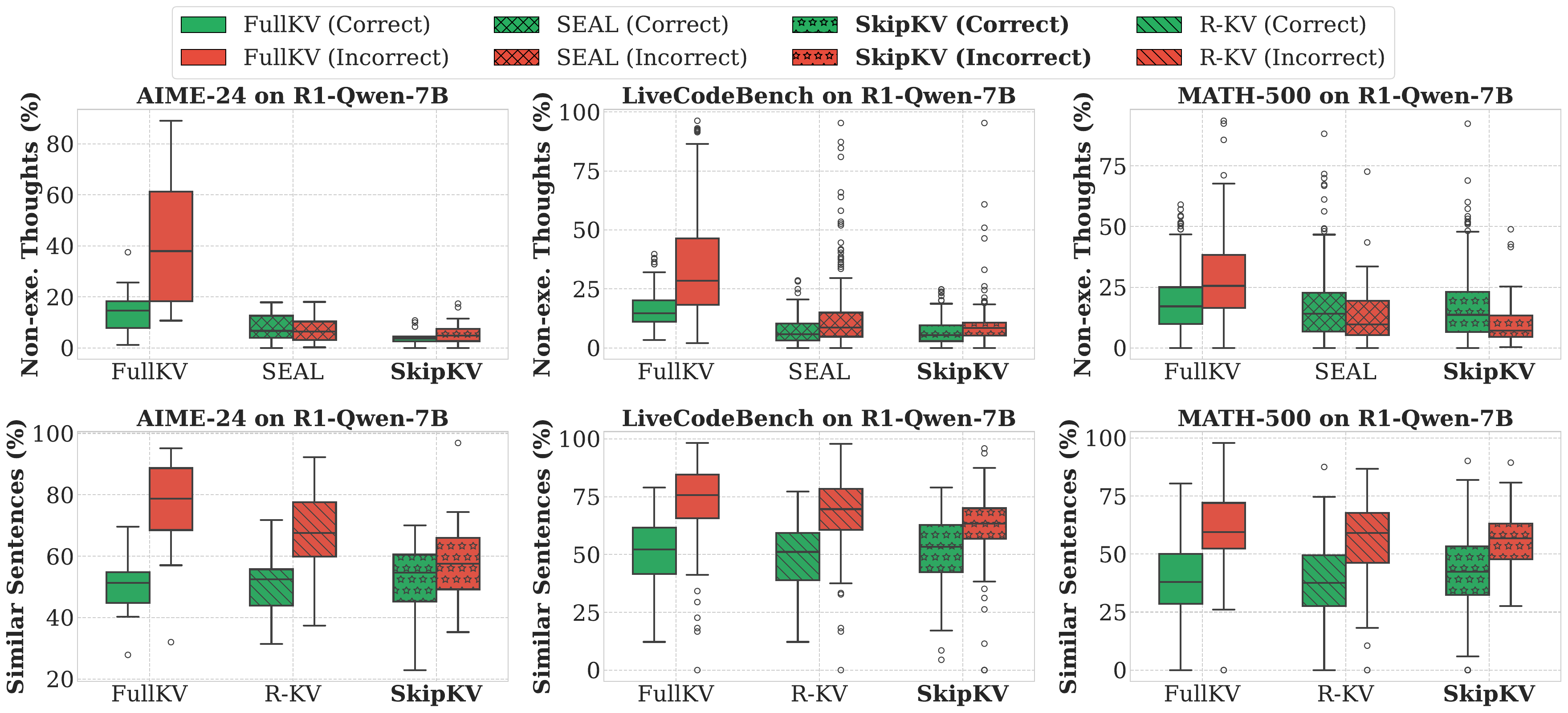}
    \caption{Comparison of the ratio of non-execution thoughts (top) and high-similarity sentences (bottom) generated by different methods on AIME-24, LiveCodeBench, and MATH-500 using R1-Qwen-7B. The boxplots show distributions for samples that were answered correctly (green) and incorrectly (red) of each method.}
    \vspace{-4mm}
    \label{fig:sentence2}
\end{figure*}

\begin{table}[t]
\centering
\small
\setlength{\tabcolsep}{6pt}
\caption{Comparison of latency and accuracy on GSM8K under different batch sizes using vLLM.}
\resizebox{0.45\textwidth}{!}{
\begin{tabular}{l|lcccc}
\toprule
\textbf{Metric} & \textbf{Batch-size} & 64 & 32 & 16 & 8 \\
\midrule
\multirow{2}{*}{\shortstack{\textbf{Total Latency} \\ (min) $\downarrow$}} 
& R-KV          & 51 & 55 & 72 & 117 \\
& \cellcolor{blue!5} SkipKV (ours) 
& \cellcolor{blue!5}\textbf{44} 
& \cellcolor{blue!5}\textbf{52} 
& \cellcolor{blue!5}\textbf{68} 
& \cellcolor{blue!5}\textbf{112} \\
\midrule
\multirow{2}{*}{\shortstack{\textbf{Accuracy} \\ (\%) $\uparrow$}}
& R-KV          & 73.8 & 81.7 & 85.3 & 87.1 \\
& \cellcolor{blue!5} SkipKV (ours) 
& \cellcolor{blue!5}\textbf{84.4} 
& \cellcolor{blue!5}\textbf{86.8} 
& \cellcolor{blue!5}\textbf{88.2} 
& \cellcolor{blue!5}\textbf{88.6} \\
\bottomrule
\end{tabular}
}
\vspace{-5mm}
\label{tab:gsm8k_kv}
\end{table}

\subsection{Additional Experimental Results}
\label{sec:appendix-main}
\paragraph{Evaluation on the Llama Model Family.}
Fig. \ref{fig:main_llama} shows the accuracy and total token length on R1-Llama-8B for MATH-500, AIME-24 and LiveCodeBench datasets comparing SkipKV with prior eviction strategies including H2O and R-KV and the FullKV baseline.
Across all three datasets, SkipKV consistently achieves the best trade-off between accuracy and generation efficiency.
In terms of accuracy, SkipKV either outperforms or matches FullKV across all KV budgets, while significantly outperforming H2O and R-KV, especially under tighter KV constraints.
In terms of efficiency, SkipKV consistently produces the shortest total generation length across all settings, indicating more concise reasoning. Notably, it reduces token length substantially compared to FullKV while simultaneously improving or preserving accuracy, demonstrating that redundant reasoning can be effectively removed without harming performance.
These results extends our evaluation beyond Qwen models in the main content, demonstrating that SkipKV generalizes effectively to other model families such as Llama.

\paragraph{System-Level Evaluation on vLLM.}
To evaluate system-level performance, we further integrate SkipKV into the vLLM v1 framework \cite{kwon2023efficient}, building upon the R-KV implementation. Unlike the HuggingFace-based PyTorch setup, which uses static batching, vLLM adopts continuous batching with a paged KV cache, eliminating explicit prefill padding within each batch. Consequently, we do not apply the batch-grouping optimization in this setting and instead focus on evaluating the core component of SkipKV—sentence-aware KV eviction and selection—under realistic inference-time scheduling and memory management.

As shown in Tab.~\ref{tab:gsm8k_kv}, under a fixed KV budget (B=512), R-KV exhibits noticeable accuracy degradation as batch size increases, indicating limited robustness under multi-request decoding. In contrast, SkipKV maintains stable accuracy across different batch sizes and achieves strong performance even at larger batch sizes, while also reducing end-to-end latency. 
These results demonstrate that SkipKV is compatible with vLLM’s paged KV cache and continuous batching mechanism, and can effectively improve both efficiency and accuracy in practical multi-request inference scenarios.

\subsection{Analysis of Sentence-level Properties after Eviction}
\label{sec:appendix-ablation}
Fig. \ref{fig:sentence2} compares the reasoning behavior of different KV eviction strategies on AIME-24, LiveCodeBench, and MATH-500 using R1-Qwen-7B.
We set the KV budget to 2220 (20\%), 2000 (28\%), 1024 (30\%) for both R-KV and SkipKV on each dataset, under which SkipKV attains the same accuracy as FullKV.
The top row of the figure illustrates the fraction of non-execution thoughts, reflecting the frequency of unnecessary re-validation steps, while the bottom row reports the proportion of high-similarity sentences, which indicate repetitive reasoning patterns.

As discussed in Observations 4 and 5 in §\ref{sec:method}, comparing correct and incorrect samples reveals that non-executable and redundant reasoning accumulates more heavily in incorrect generations. SkipKV effectively narrows this gap, reducing unnecessary and repetitive thoughts in both cases. 
Quantitatively, SkipKV generates approximately $4\times$ and $8\times$ fewer non-execution thoughts in correct and incorrect generations compared to FullKV, and further outperforms SEAL by $1.8\times$ and $1.3\times$, respectively.

Moreover, SkipKV reduces the variance of the non-execution-thought ratio across samples, enabled by its sample-wise adaptive steering mechanism. Empirically, applying an initial steering strength $\alpha_0=1.0$ reduces the number of non-execution thoughts from a wide range (up to $\sim$1000) to a controlled range (up to $\sim$50). With $\gamma=0.02$, the resulting $\alpha_t \in [1,2]$ remains within a safe regime identified in prior work~\cite{chen2025seal}, while enabling dynamic, sample-dependent adjustment based on the evolving generation.

We also observed that, although R-KV aims to remove redundant tokens, it only reduces similar sentences in incorrect answers by up to 11\%, whereas SkipKV—explicitly designed to target redundant sentences—consistently achieves over $2\times$ greater reduction.
Overall, SkipKV demonstrates a clear advantage in maintaining concise, execution-oriented reasoning under constrained KV budgets.

\subsection{Empirical Study of Generated Outputs}
\begin{figure*}[htbp]
    \centering
    \fcolorbox{blue!70!black}{white}{%
    \includegraphics[height=0.9\textheight]{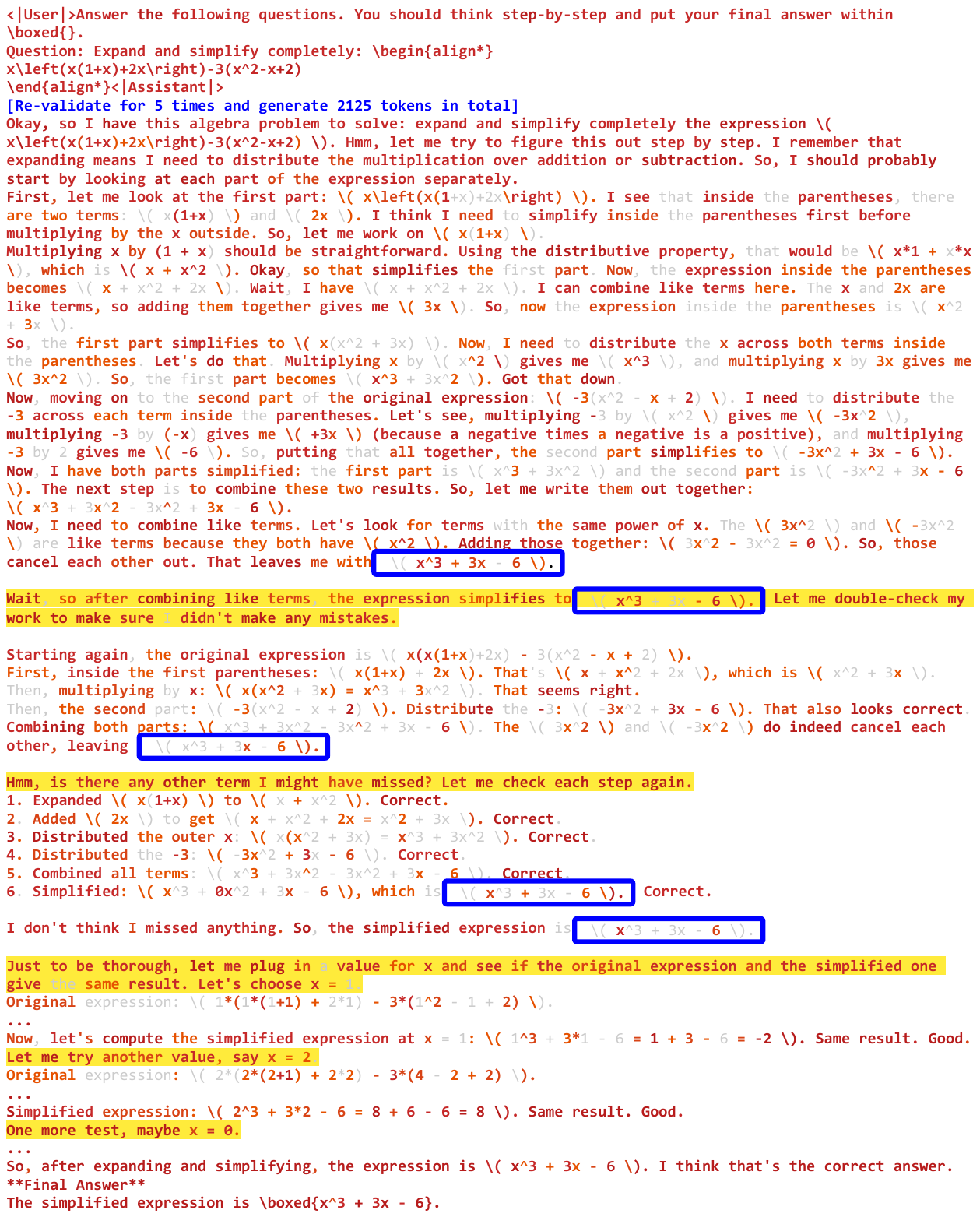}
    }
    \caption{Qualitative example of \textbf{R-KV} responses on the MATH-500 dataset.
    The darkness of \textcolor{red}{red} denotes how many heads select the token.
    Non-execution sentences starting re-validation are \colorbox{yellow}{highlighted in yellow}, where each is followed by few execution thoughts, and the answers are highlighted in a \textcolor{blue}{blue box}.
    R-KV frequently selects fragmented tokens within execution reasoning and always includes parts of the answer itself.
    }
    \label{fig:rkv_visual}
\end{figure*}

\begin{figure*}[htbp]
    \centering
    \fcolorbox{blue!70!black}{white}{%
    \includegraphics[height=0.9\textheight]{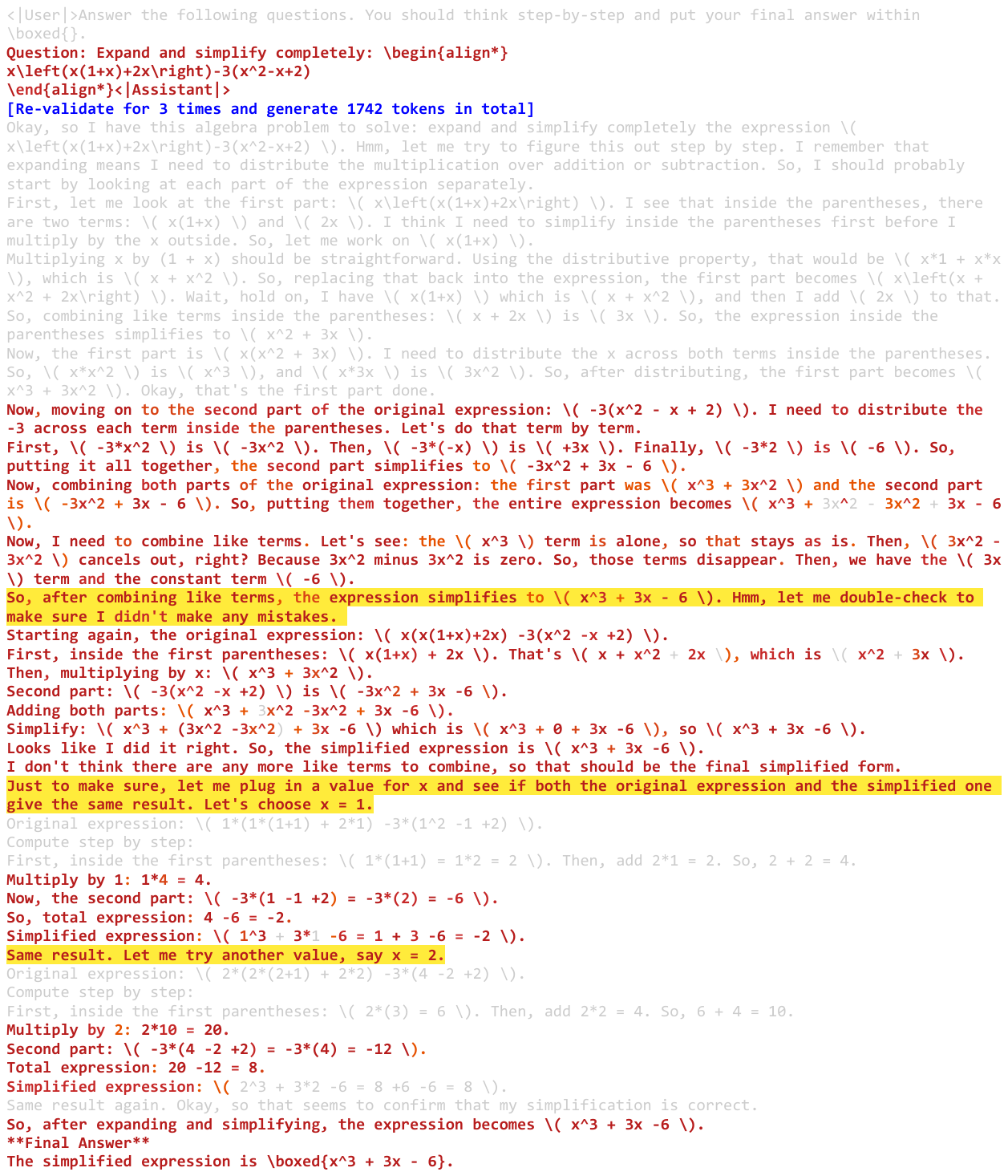}
    }
    \caption{
    Qualitative example of \textbf{SkipKV} responses on the MATH-500 dataset.
    SkipKV primarily evicts entire sentences instead of fragmented tokens, avoiding interrupting the reasoning path.
    }
    \label{fig:skipkv_visual}
\end{figure*}

In this section, we present qualitative examples from the MATH-500 dataset to compare the generation behaviors of R-KV and SkipKV (Figures~\ref{fig:rkv_visual} and~\ref{fig:skipkv_visual}).
We visualize the generated outputs of the R1-Qwen-7B reasoning model under a KV-cache budget of 1024.
In the selected example, both methods produce the correct final answer; however, SkipKV generates approximately 20\% fewer tokens than R-KV.
In both figures, non-execution thoughts are highlighted in yellow and final-answer segments are highlighted in a blue box. The darkness of the red text indicates the number of attention heads selecting each token.
From the visualization of eviction patterns, SkipKV primarily removes complete sentences from the KV cache, whereas R-KV tends to evict fragmented tokens.
The redundancy-based scoring of R-KV often leads it to remove numerical tokens that lie along crucial mathematical reasoning paths, disrupting logical consistency and resulting in longer generations.
Notably, R-KV also evicts tokens from the final-answer region, which confuses the model and triggers unnecessary re-validation steps.
In contrast, SkipKV preserves the key reasoning and answer segments, demonstrating more coherent and efficient generation behavior.


\newpage

\end{document}